\documentclass{article}

\usepackage{microtype}
\usepackage{graphicx}
\usepackage{booktabs} 

\usepackage{hyperref}

\usepackage[accepted]{icml2018}

\usepackage{url}  

\usepackage{natbib}
\usepackage{subcaption}

\usepackage{todo}
\usepackage{soul}

\usepackage{amsmath} 
\usepackage{amssymb}  

\newcommand{\eref}[1]{(\ref{#1})}
\newcommand{\aref}[1]{Algorithm~\ref{#1}}
\newcommand{\sref}[1]{Section~\ref{#1}}
\newcommand{\figref}[1]{Figure~\ref{#1}}
\newcommand{\bs}[1]{\boldsymbol{#1}}

\newcommand\tab[1][1cm]{\hspace*{#1}}

\renewcommand{\vec}[1]{\mathbf{#1}}

\newboolean{include-notes}
\setboolean{include-notes}{true}
\newcommand{\ssnote}[1]{\ifthenelse{\boolean{include-notes}}%
 {\textcolor{red}{\textbf{[ #1  --Sidd]}}}{}}
\newcommand{\dnote}[1]{\ifthenelse{\boolean{include-notes}}%
 {\textcolor{magenta}{\textbf{[ #1  --Drew]}}}{}}
 \newcommand{\gnote}[1]{\ifthenelse{\boolean{include-notes}}%
 {\textcolor{purple}{\textbf{[ #1 --Geoff]}}}{}}
  \newcommand{\ah}[1]{\ifthenelse{\boolean{include-notes}}%
 {\textcolor{green}{\textbf{[ #1  --Ahmed]}}}{}}
 \newcommand{\wen}[1]{\ifthenelse{\boolean{include-notes}}%
 {\textcolor{blue}{\textbf{[#1  --Wen]}}}{}}
  \newcommand{\zm}[1]{\ifthenelse{\boolean{include-notes}}%
 {\textcolor{cyan}{\textbf{[ #1  --Zita]}}}{}}
\newcommand{\add}[1]{\ifthenelse{\boolean{include-notes}}%
 {\textcolor{blue}{#1}}{#1}}

\usepackage{enumitem}
\begin{document} 

\twocolumn[
\icmltitle{Recurrent Predictive State Policy Networks}

\icmlsetsymbol{equal}{*}

\begin{icmlauthorlist}
\icmlauthor{Ahmed Hefny}{equal,mld}
\icmlauthor{Zita Marinho}{equal,ri,ist}
\icmlauthor{Wen Sun}{ri}
\icmlauthor{Siddhartha Srinivasa}{wa}
\icmlauthor{Geoffrey Gordon}{mld}
\end{icmlauthorlist}

\icmlaffiliation{mld}{Machine Learning Department, Carnegie Mellon University, Pittsburgh, USA}
\icmlaffiliation{ri}{Robotics Institute, Carnegie Mellon University, Pittsburgh, USA}
\icmlaffiliation{ist}{ISR/IT, Instituto Superior T\'{e}cnico, Lisbon, Portugal}
\icmlaffiliation{wa}{Paul G. Allen School of Computer Science \& Engineering, University of Washington, Seattle, USA}

\icmlcorrespondingauthor{Ahmed Hefny}{ahefny@cs.cmu.edu}
\icmlcorrespondingauthor{Zita Marinho}{zmarinho@cmu.edu}

\vskip 0.3in
]

\printAffiliationsAndNotice{\icmlEqualContribution}

\begin{abstract} 
We introduce Recurrent Predictive State Policy (RPSP) networks,
a recurrent architecture that brings insights from predictive state representations to reinforcement learning in partially observable environments. 
Predictive state policy networks consist of a recursive filter, which keeps track of a belief about the state of the environment, and a reactive policy that directly maps beliefs to actions, to maximize the cumulative reward. 
The recursive filter leverages predictive state representations (PSRs) \citep{tpsr, psim} by modeling \emph{predictive state} --- a prediction of the distribution of future observations conditioned on history and future actions. This representation gives rise to a rich class of statistically consistent algorithms \citep{hefny:17} to initialize the recursive filter. Predictive state serves as an equivalent representation of a belief state. Therefore, the policy component of the RPSP-network can be purely reactive, simplifying training while still allowing optimal behaviour. 
Moreover, we use the PSR interpretation during training as well,
by incorporating prediction error in the loss function.
The entire network (recursive filter and reactive policy) is still differentiable and can be trained using gradient based methods. 
 We optimize our policy using a combination of policy gradient based on rewards~\citep{Williams92} and gradient descent based on prediction error.  
 We show the efficacy of RPSP-networks under partial observability on a set of robotic control tasks from OpenAI Gym. We empirically show that RPSP-networks perform well compared with memory-preserving networks such as GRUs, as well as finite memory models, being the overall best performing method.  
\end{abstract} 

\section{Introduction}
\label{sec:intro}

Recently, there has been significant progress in deep reinforcement learning~\citep{Bojarski16,schulman15,mnih:13:atari,Silver16}.
Deep reinforcement learning combines deep networks as a high level representation of a policy with reinforcement learning as an optimization method, and allows for end-to-end training.

While traditional applications of deep learning rely on standard architectures with sigmoid activations or rectified linear units, there is an emerging trend of using composite architectures that contain 
parts explicitly resembling other algorithms such as Kalman filtering~\citep{bpkf}
and value iteration~\citep{vin}. It has been shown that such architectures can outper,form standard neural networks.

In this work, we focus on partially observable environments, where the agent 
does not have full access to the state of the environment, but only to partial observations thereof.
The agent has to maintain instead a distribution over states, \emph{i.e.}, a belief state, based on the entire history 
of observations and actions. 
The standard approach to this problem is to employ recurrent architectures such as Long-Short-Term-Memory (LSTM)~\citep{Hochreiter97} and Gated Recurrent Units (GRU)~\citep{Cho14}. 
However, these recurrent models are difficult to train due to non-convexity, and their hidden states lack a statistical meaning and are hard to interpret. 

Models based on predictive state representations~\citep{Littman01,psr,tpsr,hsepsr} 
offer an alternative method to construct a surrogate for belief state in a partially observable environment. These models represent state as the expectation of sufficient statistics of future observations, conditioned on  history and future actions. Predictive state models admit efficient learning algorithms with theoretical guarantees. 
Moreover, the successive application of the predictive state update procedure (i.e., filtering equations) results in a recursive computation graph that is differentiable with respect to model parameters. Therefore, we can treat predictive state models as recurrent networks and apply backpropagation through time (BPTT) \citep{hefny:17,downey:17} to optimize model parameters. We use this insight to construct a \emph{Recurrent Predictive State Policy} (RPSP) network, a special recurrent  architecture that consists of (1) a predictive state model acting as a recursive filter to keep track of a predictive state,
and (2) a feed-forward neural network that directly maps predictive states to actions.
This configuration results in a recurrent policy, where the recurrent part is implemented by a PSR instead of an LSTM or a GRU.
As predictive states are a sufficient summary of the history of observations and actions, the reactive policy will have rich enough information to make its decisions, as if it had access to a true belief state.
 There are a number of motivations for this architecture:
\begin{itemize}
    \item Using a PSR means we can benefit from methods in the spectral learning literature to provide an efficient and statistically consistent initialization of a core component of the policy.
    \item Predictive states have a well defined statistical interpretation and hence they can be examined and optimized based on that interpretation. 
    \item The recursive filter in RPSP-networks is fully differentiable, meaning that once a good initialization is obtained from spectral learning methods, we can refine RPSP-nets using gradient descent. 
\end{itemize}

This network can be trained end-to-end, for example using policy gradients in a reinforcement learning setting~\citep{pgrad} or supervised learning in an imitation learning setting~\citep{dagger}. In this work we focus on the former. 
We discuss the predictive state model component in \S\ref{sec:psr}.
The control component is presented in \S\ref{sec:policynetwork} 
and the learning algorithm is presented in \S\ref{sec:policygradient}.
In \S\ref{sec:experiments} we describe the experimental setup and results on control tasks: we evaluate the performance of reinforcement learning using predictive state policy networks in multiple
partially observable environments with continuous observations and actions.


\section{Background and Related Work}
Throughout the rest of the paper, we will define vectors in bold notation $\vec{v}$, matrices in capital letters $W$. We will use $\otimes$ to denote vectorized outer product: $\vec{x} \otimes \vec{y}$ is $\vec{x} \vec{y}^\top$ reshaped into a vector.

We assume an agent is interacting with the environment in episodes, 
where each episode consists of $T$ time steps in each of which the agent 
takes an action $\vec{a}_t \in \mathcal{A}$, and observes an observation $\vec{o}_t  \in \mathcal{O}$ and a reward $r_t \in \mathbb{R}$.
The agent chooses actions based on a stochastic policy $\pi_{\bs\theta}$ parameterized 
by a parameter vector $\theta$:
$\pi_{\bs\theta}(\vec{a}_t \mid \vec{o}_{1:t-1},\vec{a}_{1:t-1}) \equiv p(\vec{a}_t \mid \vec{o}_{1:t-1},\vec{a}_{1:t-1}, \bs\theta)$.
We would like to improve the policy rewards by optimizing $\bs\theta$ based on the agent's experience in order to maximize the expected long term reward
$J(\pi_{\bs\theta}) = \frac{1}{T} \sum_{t=1}^T \mathbb{E}\left[\gamma^{t-1}r_t \mid \pi_{\bs\theta}\right]$, where $\gamma\in[0,1]$ is a discount factor.

There are two major approaches for policy modeling and optimization.
The first is the value function-based approach,
where we seek to learn a function (e.g., a deep network~\citep{mnih:13:atari}) to evaluate the value of each action at each state (a.k.a. Q-value) under the optimal policy~\citep{Sutton98}. Given the Q function the agent can act greedily based on estimated values.
The second approach is direct policy optimization, where we learn a function to directly predict optimal actions (or optimal action distributions). This function is optimized to maximize $J(\theta)$ using policy gradient methods \citep{schulman15,Duan16} or derivative-free methods~\citep{cem}.
We focus on the direct policy optimization approach as it is more robust to noisy continuous environments and modeling uncertainty~\citep{pgrad,WierstraRPG}. 

Our aim is to provide a new class of policy functions that combines recurrent reinforcement learning with recent advances in modeling partially observable environments using predictive state representations (PSRs). 
There have been previous attempts to combine predictive state models with policy learning. \citet{boots:11:plan} proposed a method for planning in partially observable environments.
The method first learns a PSR from a set of trajectories collected using an explorative blind policy.
The predictive states estimated by the PSR are then considered as states in a fully observable Markov Decision Process. A value function is learned on these states using
least squares temporal difference~\citep{PSTD} or
point-based value iteration (PBVI)~\citep{boots:11:plan}.
The main disadvantage of these approaches is that it
assumes a one-time initialization of the PSR and does not propose a mechanism
to update the model based on subsequent experience.

\citet{Hamilton14} proposed an iterative method to simultaneously learn a PSR and use the predictive states to fit a Q-function. \citet{specpomdp} proposed a tensor decomposition method to estimate 
the parameters of a discrete partially observable Markov decision process (POMDP).
One common limitation in the aforementioned methods is that they are restricted to 
discrete actions (some even assume discrete observations).
Also, it has been shown that PSRs can benefit greatly from local optimization
after a moment-based initialization \citep{downey:17,hefny:17}.

\citet{psd} proposed predictive state decoders, where an LSTM or a GRU network is trained on a mixed objective function in order to obtain high cumulative rewards while accurately predicting future observations.
While it has shown improvement over using standard training objective functions, it does not solve the initialization issue of
the recurrent network.

Our proposed RPSP networks alleviate the limitations of previous approaches: It supports continuous observations and actions, it uses a recurrent state tracker with consistent initialization, and it supports end-to-end training after the initialization. 

\section{Predictive State Representations of Controlled Models}
\label{sec:psr}

\begin{figure*}[t!]
\centering
\begin{subfigure}[t]{0.25\textwidth}
\includegraphics[page=2,clip,trim={0cm 2.5cm 20cm 0cm},scale=0.6]{figs/sketch/figs.pdf}
\label{fig:RNNPSR}
\end{subfigure}
\hfill
\begin{subfigure}[t]{0.55\textwidth}
\includegraphics[page=1,width=\textwidth,clip,trim={4cm 5.9cm 16cm 0.5cm},scale=0.8]{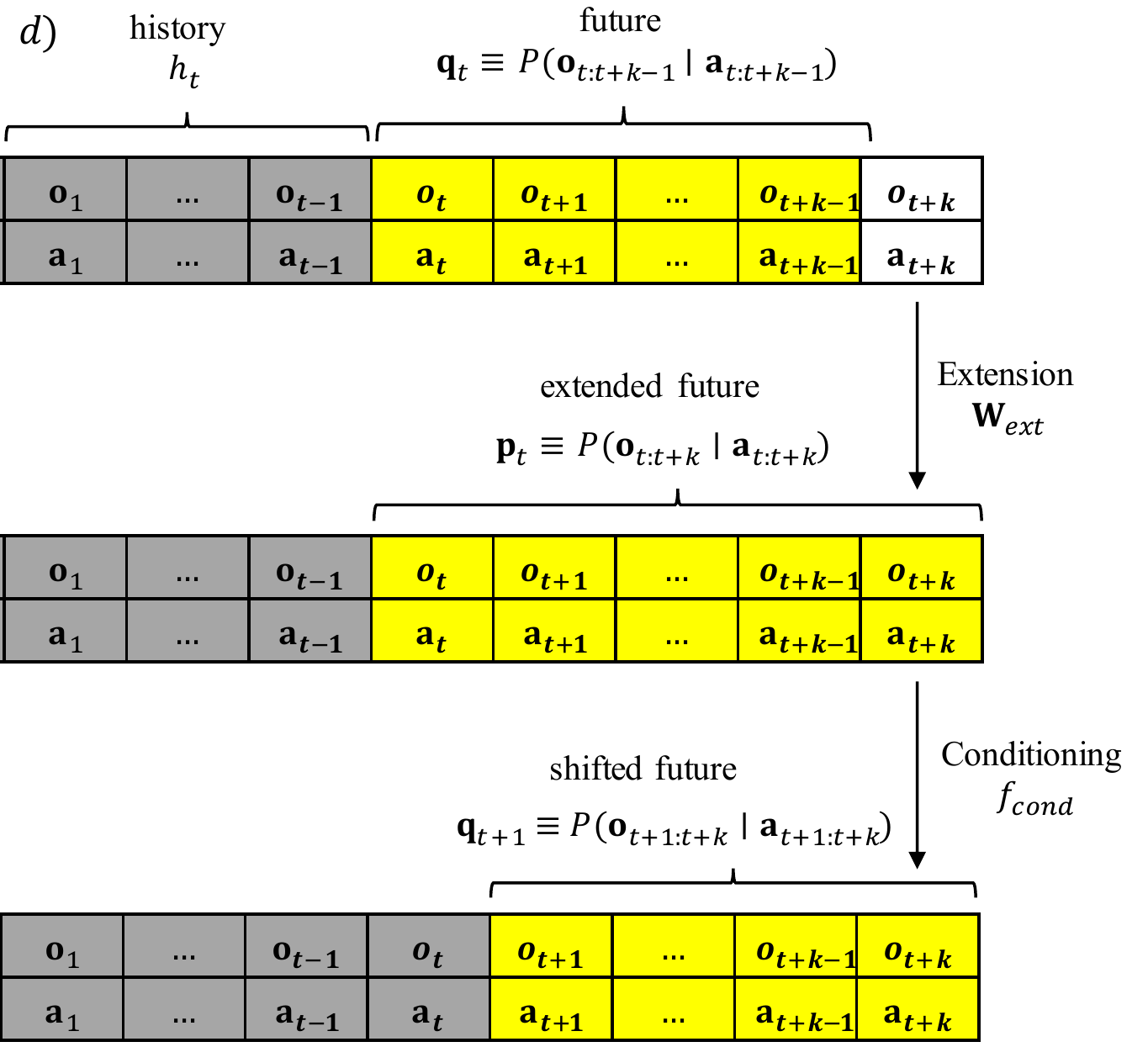}
\end{subfigure}
\caption{
\textbf{Left:} a) Computational graph of RNN and PSR
and b) the details of the state update function $f$ for both a simple RNN 
and c) a PSR.
Compared to RNN, the observation function $g$ is easier to learn in a PSR
(see \S\ref{sec:pscm}).
\textbf{Right:} 
Illustration of the PSR extension and conditioning steps.\vspace{-0.4cm}}
\label{fig:windows}
\end{figure*}

In this section, we give a brief introduction to predictive state representations, which constitute the state tracking (filtering)
component of our model and we discuss their relationship to recurrent neural networks RNNs.\footnote{
We follow the predictive state controlled model formulation in \citet{hefny:17}. 
However, alternative methods such as predictive state inference machines \citep{psim} could be contemplated.}
Given a history of observations and actions $\vec{a}_1,\vec{o}_1,\vec{a}_2,\vec{o}_2,\dots,\vec{a}_{t-1},\vec{o}_{t-1}$,
a typical RNN computes a hidden state $\vec{q}_t$ using a recursive update equation $\vec{q}_{t+1} = f(\vec{q}_t,\vec{a}_t,\vec{o}_t)$. 
Given the state $\vec{q}_t$, one can predict observations through a function $g(\vec{q}_t,\vec{a}_t) \equiv \mathbb{E}[\vec{o}_t \mid \vec{q}_t, \vec{a}_t]$.
Because $\vec{q}$ is latent, the function $g$ that connects states to the output is unknown
and has to be learned. In this case, the output could be predicted from observations, when the RNN is used for prediction, see \figref{fig:windows} (a,b).

Predictive state models define a similar recursive state update. 
However, the state $\vec{q}_t$ has a specific interpretation: it encodes a conditional distribution
of future observations
$\vec{o}_{t:t+k-1}$ conditioned on future actions $\vec{a}_{t:t+k-1}$.\footnote{The length-$k$ depends on the observability of the system. A system is $k$-observable if maintaining the predictive state is equivalent to maintaining
the distribution of the system's latent state.}
(for example, in the discrete case, $\vec{q}_t$ could be a vectorized probability table).
We denote this representation of state as \emph{predictive state}.

The main characteristic of a predictive state is that it is defined entirely
in terms of observable quantities. That is, the mapping between the predictive state $\vec{q}_t$ and the prediction of $\vec{o}_t$ given $\vec{a}_t$ is implied by the choice of features.
With suitable features, the mapping can be fully known or simple to learn consistently.
This is in contrast to a general RNN, where this mapping is unknown and typically requires non-convex optimization to be learned.
This characteristic allows for efficient and consistent learning  of models with predictive states 
by reduction to supervised learning~\citep{ivr,psim}.\footnote{
We do not need to be interested in prediction to take advantage 
of PSR initialization.
}

Similar to an RNN, a PSR employs a recursive state update that consists of the following two steps:\footnote{See the appendix~\sref{sec:ivr} for more details.}
\begin{itemize}
    \item State extension: A linear map $W_\mathrm{ext}$ is applied to $\vec{q}_t$ to obtain an \emph{extended state} $\vec{p}_t$. This state defines a conditional distribution over an extended window of $k+1$ observations and actions. 
    $W_\mathrm{ext}$ is a parameter to be learned.
    \begin{align}\label{eq:linear}
    \vec{p}_t = W_\mathrm{ext}\vec{q}_t
    \end{align}
    \item Conditioning: Given $\vec{a}_t$ and $\vec{o}_t$, and a known conditioning function $f_\mathrm{cond}$: 
    \begin{align}
    \vec{q}_{t+1} = f_\mathrm{cond}(\vec{p}_t,\vec{a}_t,\vec{o}_t).
    \label{eq:filter_cond}
    \end{align}
\end{itemize}

Figure \ref{fig:windows} (c, d) depicts the PSR state update.
The conditioning function $f_{\mathrm{cond}}$ depends on the representation of $\vec{q}_t$ and $\vec{p}_t$.
For example, in a discrete system, $\vec{q}_t$ and $\vec{p}_t$ could represent conditional probability tables and $f_\mathrm{cond}$ amounts to applying Bayes 
rule. Extension to continuous systems is achieved using Hilbert space embedding of distributions \citep{hsepsr}, where
$f_\mathrm{cond}$ uses kernel Bayes rule \citep{fukumizu13}.

In this work, we use the RFFPSR model proposed in~\citep{hefny:17}.
Observation and action features are based on random Fourier features (RFFs) of RBF kernel \citep{rff}
projected into a lower dimensional subspace using randomized PCA \citep{halko:11:randsvd}.
We use $\phi$ to denote this feature function.
Conditioning function $f_{cond}$ is kernel Bayes rule,
and observation function $g$ is a linear function of state
$\mathbb{E}[\vec{o}_t \mid \vec{q}_t, \vec{a}_t] = {W}_\mathrm{pred} (\vec{q}_t \otimes \phi(\vec{a}_t))$. 

\subsection{Learning predictive states representations}
\label{sec:pscm}
Learning PSRs is carried out in two steps: an initialization procedure using method of moments
and a local optimization procedure using gradient descent.

\textbf{Initialization:  } The initialization procedure exploits the fact that $\vec{q}_t$ and $\vec{p}_t$ are represented in terms of observable quantities: since $W_\mathrm{ext}$ is linear and using \eref{eq:linear}, then $\mathbb{E} [\vec{p}_t \mid \vec{h}_t] = W_\mathrm{ext} \mathbb{E} [ \vec{q}_t \mid \vec{h}_t]$. Here $\vec{h}_t \equiv h(\vec{a}_{1:t-1},\vec{o}_{1:t-1})$ denotes a set of features extracted from previous observations and actions (typically from a fixed length window ending at $t-1$). Because $\vec{q}_t$ and $\vec{p}_t$ are not hidden states, estimating 
these expectations on both sides can be done by solving a supervised regression subproblem. Given the predictions from this regression, solving for $W_\mathrm{ext}$ then becomes another linear regression problem. Here, we follow this two-stage regression proposed by ~\citet{ivr}.\footnote{
Specifically, we use the joint stage-1 regression variant for initialization.}
Once $W_\mathrm{ext}$ is computed, we can perform filtering to obtain the predictive states $\vec{q}_t$. We then use the estimated states  
to learn the mapping to predicted observations ${W}_\mathrm{pred}$, which results in another regression subproblem, see Section \ref{sec:ivr} for a detailed derivation.

In RFFPSR, we use linear regression for all subproblems (which is a reasonable choice with kernel-based features).
This ensures that the two-stage regression procedure is free of local optima. 

\textbf{Local Optimization:  }
Although PSR initialization procedure is consistent, 
it is based on method of moments and hence is not necessarily statistically efficient.
Therefore it can benefit from local optimization.
\citet{downey:17} and \citet{hefny:17} note that a PSR defines a recursive computation graph similar to that of an RNN where we have
\begin{align}
\nonumber \vec{q}_{t+1} & = f_\mathrm{cond}(W_\mathrm{ext}(\vec{q}_t),\vec{a}_t,\vec{o}_t)) \\
\mathbb{E}[\vec{o}_t \mid \vec{q}_t, \vec{a}_t] & = {W}_\mathrm{pred} (\vec{q}_t \otimes \phi(\vec{a}_t)),
\label{eq:psr_filter}
\end{align}
With a differentiable $f_\mathrm{cond}$, the PSR can be trained using backpropagation through time \citep{bptt}
to minimize prediction error. 

In a nutshell, a PSR effectively constitutes a special type of a recurrent network where the state representation and update 
are chosen in a way that permits a consistent initialization, which is then followed by conventional backpropagation.

\section{Recurrent Predictive State Policy (RPSP) Networks}
\label{sec:policynetwork}
\begin{figure}[t!]
\centering
\includegraphics[page=3,width=\columnwidth,clip,trim={0cm 7cm 20cm 0cm},scale=0.6]{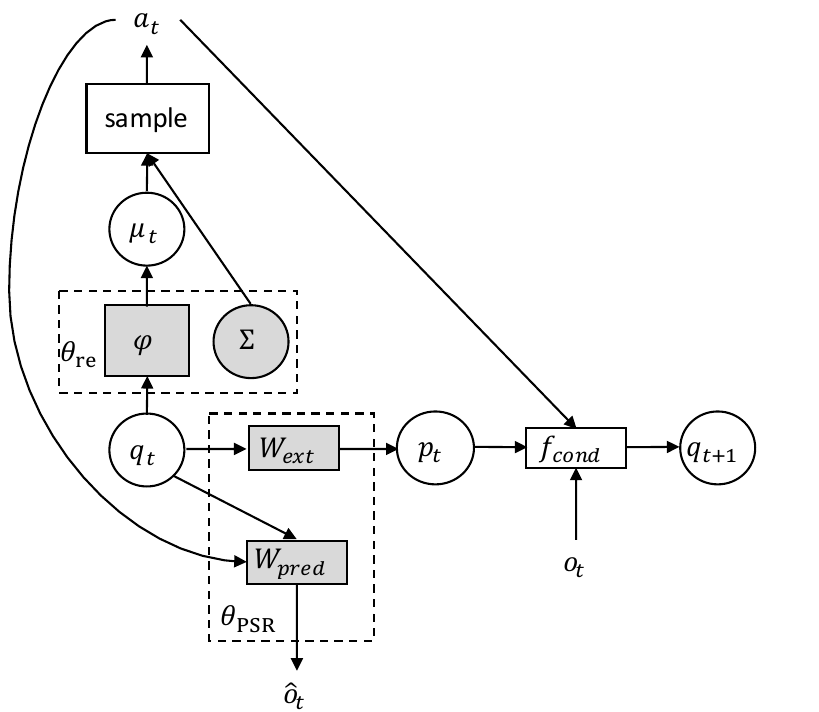}
\caption{RPSP network: The predictive state is updated 
by a linear extension $W_\mathrm{ext}$ followed by a non-linear conditioning $f_\mathrm{cond}$. 
A linear predictor $W_\mathrm{pred}$ is used to predict observations, which is used to regularize training loss (see \S\ref{sec:policygradient}).
A feed-forward reactive policy maps the predictive states $\vec{q}_t$
to a distribution over actions.}
\label{fig:RPN}
\end{figure}

We now introduce our proposed class of policies, Recurrent Predictive State Policies (RPSPs). 
In this section, we formally describe its components and we describe
the policy learning algorithm in~\S\ref{sec:policygradient}.

RPSPs consist of two fundamental components: a state tracking component, which models the state of the system, and is able to predict future observations; and a reactive policy component, that maps states to actions, shown in \figref{fig:RPN}.

The state tracking component is based on the PSR formulation described in \S\ref{sec:psr}.
For the reactive policy, we consider a stochastic non-linear policy $\pi_\mathrm{re}(\vec{a}_t \mid \vec{q}_t) \equiv p(\vec{a}_t \mid \vec{q}_t;\boldsymbol\theta_\mathrm{re})$ 
which maps a predictive state to a distribution over actions and is parametrized by $\bs{\theta}_\mathrm{re}$.
Similar to \citet{schulman15} we assume a Gaussian distribution ${\cal N}(\bs{\mu}_t, {\Sigma})$, with parameters 
\begin{align}
\label{eq:policy} &
\boldsymbol{\mu} = \varphi(\vec{q}_t; \boldsymbol{\theta}_{\mu}); &&
{\Sigma} = \mathrm{diag}(\exp(\vec{r}))^{2}
\end{align}
and a non-linear map $\varphi$ parametrized by $\boldsymbol{\theta}_{\mu}$ (e.g. a feedforward network)
, and a learnable vector $\vec{r}$.

An RPSP is thus a stochastic recurrent policy with the recurrent part corresponding to a PSR.
The parameters $\boldsymbol{\theta}$ consist of two parts:  
the PSR parameters $\boldsymbol\theta_\mathrm{PSR}=\{\vec{q}_0,W_\mathrm{ext},W_\mathrm{pred}\}$
and the reactive policy parameters
$\boldsymbol\theta_\mathrm{re}=\{\boldsymbol\theta_\mu,\vec{r}\}$.
In the following section, we describe how these parameters are learned.

\section{Learning RPSPs}
\label{sec:policygradient}

\begin{algorithm}[t]
   \caption{Recurrent Predictive State Policy network Optimization (RPSPO)}
   \label{alg:psrnetwork}
\small
\begin{algorithmic}[1]
\STATE \textbf{Input:} Learning rate $\eta$.
\STATE Sample initial trajectories: $\{(o^i_t,a^i_t)_{t}\}_{i=1}^{M}$ from $\pi_{exp}$.
\STATE Initialize PSR: \par
$\ \ \ \ \ {\boldsymbol{\theta}^0_{\mathrm{PSR}}}=\{\vec{q}_0, W_\text{ext},W_\text{pred}\}$ via 2-stage regression in~\S\ref{sec:psr}.
\STATE Initialize reactive policy $\boldsymbol{\theta}_{\mathrm{re}}^0$ randomly.
\FOR{$n=1\dots N_{max}$ iterations}
\FOR{$i=1,\dots,M$ batch of $M$ trajectories from $\pi^{n-1}$:}
\STATE Reset episode: $a^i_0$.
\FOR{$t=0\dots T$ \emph{roll-in} in each trajectory:}
\STATE Get observation $o^i_t$ and reward $r^i_t$.
\STATE Filter $\vec{q}^i_{t+1}=f_t(\vec{q}^i_{t},\vec{a}^i_{t},\vec{o}^i_{t})$ in (Eq.~\ref{eq:psr_filter}).
\STATE Execute $\vec{a}^i_{t+1}\sim\pi^{n-1}_{re}(\vec{q}^i_{t+1})$.
\ENDFOR
\ENDFOR
\STATE Update $\boldsymbol{\theta}$ using $\mathcal{D}=\{\{\vec{o}^i_t,\vec{a}^i_t,r^i_t,\vec{q}_t^i\}_{t=1}^T\}_{i=1}^{M}$:\par 
$\ \ \ \ \ \ \boldsymbol{\theta}^{n} \leftarrow \textsc{Update}(\boldsymbol{\theta}^{n-1}, \mathcal{D},\eta)$, as in \S\ref{sec:policygradient}.
\ENDFOR
\STATE \textbf{Output:} Return $\boldsymbol{\theta}=(\boldsymbol{\theta}_\mathrm{PSR}, \boldsymbol{\theta}_\mathrm{re})$.
\end{algorithmic}
\end{algorithm}

As detailed in \aref{alg:psrnetwork}, learning an RPSP is performed in two phases.\footnote{We will provide a link to the implementation here.}
In the first phase, we execute an exploration policy to collect a dataset that is used to initialize the PSR
as described in \S\ref{sec:pscm}.
It is worth noting that this initialization procedure depends on observations rather than rewards.
This can be particularly useful in environments where informative reward signals are infrequent.

In the second phase, starting from the initial PSR and a random reactive policy,
we iteratively collect trajectories using the current policy and use them to update
the parameters of both the reactive policy $\boldsymbol{\theta}_\mathrm{re}=\{\boldsymbol{\theta}_{\mu}, \vec{r}\}$
and the predictive model $\boldsymbol{\theta}_\mathrm{PSR}=\{\vec{q}_0,W_\mathrm{ext},W_\mathrm{pred}\}$, as depicted in Algorithm \ref{alg:psrnetwork}.
Let $p(\tau \mid \boldsymbol{\theta})$ be the distribution over trajectories induced by the policy 
$\pi_{\boldsymbol{\theta}}$.
By updating parameters, we seek to minimize the objective function in \eref{eq:losspsr}.
\begin{align}\label{eq:losspsr}
&\mathcal{L}(\theta) = \alpha_1 \ell_1(\theta) + \alpha_2 \ell_2(\theta)\\ \notag
&=-\alpha_1 J(\pi_{\theta})+ \alpha_2 \sum_{t=0}^{T} \mathbb{E}_{p(\tau|\bs\theta)}\left[\| W_{pred} ({\vec{q}}_t \otimes \vec{a}_{t})- \vec{o}_{t}\|^2 \right] ,
\end{align}
which combines negative expected returns with PSR prediction error.\footnote{We minimize 1-step prediction error, as opposed to general $k$-future prediction error recommended by \citep{hefny:17}, to avoid biased estimates induced by non causal statistical correlations (observations correlated with future actions) when performing on-policy updates when a non-blind policy is in use.}
Optimizing the PSR parameters to maintain low prediction error can be thought of as a regularization scheme. 
The hyper-parameters $\alpha_1,\alpha_2\in\mathbb{R}$ determine the importance of the expected return and prediction error respectively. They are discussed in more detail in \S\ref{sec:normalization}.

Noting that RPSP is a special type of a recurrent network policy, it is possible to 
adapt policy gradient methods \cite{Williams92} to the joint loss in \eqref{eq:losspsr}.
In the following subsections, we propose different update variants.

\subsection{Joint Variance Reduced Policy Gradient (VRPG)}
\label{sec:joint}
In this variant, we use REINFORCE method~\citep{Williams92} to
obtain a stochastic gradient of $J(\pi)$ from a batch of $M$ trajectories.

Let $R(\tau) = \sum_{t=1}^T \gamma^{t-1} r_t$ be the cumulative discounted reward 
for trajectory $\tau$ given a discount factor $\gamma \in [0,1]$.
REINFORCE uses the likelihood ratio trick $\nabla_{\boldsymbol{\theta}} p(\tau | \boldsymbol{\theta}) = p(\tau | \boldsymbol{\theta}) \nabla_{\boldsymbol{\theta}} \log p(\tau | \boldsymbol{\theta})$
to compute $\nabla_{\boldsymbol{\theta}} J(\pi)$ as
\begin{align*}
    \nabla_{\boldsymbol{\theta}} J(\pi) = \mathbb{E}_{\tau \sim p(\tau | \boldsymbol{\theta})}[R(\tau) \sum_{t=1}^T \nabla_{\boldsymbol{\theta}} \log \pi_{\boldsymbol{\theta}}(\vec{a}_t|\vec{q}_t)],
\end{align*}

In practice, we use a variance reducing variant of policy gradient~\citep{Greensmith:2001} given by
\begin{align}
\label{eq:reinf}
\nabla_{\boldsymbol{\theta}}J(\pi) &= \mathbb{E}_{ \tau \sim p(\tau|\boldsymbol{\theta})} \sum_{t=0}^{T} [\nabla_{\boldsymbol{\theta}}\log \pi_{\boldsymbol{\theta}}(\vec{a}_t|\vec{q}_t) (R_t(\tau)-b_t)],
\end{align}
where we replace the cumulative trajectory reward $R(\tau)$ by a 
reward-to-go function $R_t(\tau) = \sum_{j=t}^T \gamma^{j-t} r_j$ computing the cumulative reward starting from $t$. To further reduce variance we use a \emph{baseline}
$b_t \equiv \mathbb{E}_{\boldsymbol{\theta}} [R_t(\tau) \mid \vec{a}_{1:t-1}, \vec{o}_{1:t}]$
which estimates the expected reward-to-go conditioned on the current policy. In our implementation, we assume $b_t = \vec{w}_b^\top \vec{q}_t$ for a parameter
vector $\vec{w}_b$ that is estimated using linear regression. 
Given a batch of $M$ trajectories,
a stochastic gradient of $J(\pi)$ can be obtained by replacing the expectation in 
\eqref{eq:reinf} with the empirical expectation over trajectories in the batch.

A stochastic gradient of the prediction error can be obtained using backpropagation through time (BPTT)~\citep{bptt}. With an estimate of both gradients, we can compute $\eqref{eq:losspsr}$ and update the parameters
trough gradient descent, see \aref{alg:vrpg} in the appendix.

\subsection{Alternating Optimization}
\label{sec:alt}

In this section, we describe a method that utilizes the recently proposed Trust Region Policy Optimization (TRPO~\citep{schulman15}), an alternative to the vanilla
policy gradient methods that has shown superior performance in practice~\citep{Duan16}. It uses a natural gradient update and enforces a constraint that encourages small changes in the policy in each TRPO step. This constraint results in smoother changes of policy parameters.

Each TRPO update is an approximate solution to the following constrained optimization problem in \eref{eq:reactiveup}.
\vspace{-0.3cm}
\begin{align}\notag
&\bs{\boldsymbol{\theta}}^{n+1} = \mathrm{arg}\min\limits_{\boldsymbol{\theta}} \mathbb{E}_{\tau \sim p(\tau|\pi^n)} \sum_{t=0}^T \left[ \frac{\pi_{\boldsymbol{\theta}}(\vec{a}_t|\vec{q}_t)}{\pi^n(\vec{a}_t|\vec{q}_t)} (R_t(\tau)-b_t) \right]  
\\
&\text{s.t. } \mathbb{E}_{\tau \sim p(\tau|\pi^n)} \sum_{t=0}^T \left[D_{KL}\left(\pi^n (.|\vec{q}_t)\mid\pi_{\boldsymbol{\theta}}(.|\vec{q}_t)\right)\right] \leq \epsilon,
\label{eq:reactiveup}
\end{align}
\vspace{-0.08cm}
where $\pi^n$ is the policy induced by $\boldsymbol{\theta}^n$, and $R_t$ and $b_t$
are the reward-to-go and baseline functions defined in \S\ref{sec:joint}.

While it is possible to extend TRPO to the joint loss in \eqref{eq:losspsr},
we observed that TRPO tends to be computationally intensive with recurrent architectures.
Instead, we resort to the following alternating optimization:
In each iteration, we use TRPO to update the reactive policy parameters $\boldsymbol{\theta}_{\mathrm{re}}$,
which involve only a feedforward network.
Then, we use a gradient step on \eqref{eq:losspsr}, as described in \S\ref{sec:joint},
to update the PSR parameters $\boldsymbol{\theta}_{\mathrm{PSR}}$, see \aref{alg:altop} in the appendix.

\subsection{Variance Normalization}
\label{sec:normalization}
It is difficult to make sense of the values of $\alpha_1$, $\alpha_2$, specially if the gradient magnitudes of their respective losses are not comparable. For this reason, we propose a more principled approach for finding the relative weights.
We use $\alpha_1 = \tilde{\alpha}_1$ and $\alpha_2 = a_2 \tilde{\alpha}_2$,
where $a_2$ is a user-given value, and $\tilde{\alpha}_1$ and $\tilde{\alpha}_2$ are dynamically adjusted to maintain the property that the gradient of each loss weighted by $\tilde{\alpha}$
has unit (uncentered) variance, in \eref{eq:normalization}. In doing so, we maintain the variance of the gradient of each loss through exponential averaging and use it to adjust the weights. 
\vspace{-0.1cm}
\begin{align}\label{eq:normalization}
\vec{v}^{(n)}_i &= (1-\beta) \vec{v}^{(n-1)}_i + \beta \sum_{\boldsymbol{\theta}_j\in \boldsymbol{\theta}} \| \nabla^{(n)}_{\boldsymbol{\theta}_{j}} \ell_i \| ^2\\\notag
\tilde{\alpha}_i^{(n)} &= \left({{\sum_{\boldsymbol{\theta}_j\in \boldsymbol{\theta}} \vec{v}_{i,j}^{(n)}}}\right)^{-1/2},
\end{align} 
\vspace{-0.9cm}

\begin{figure*}[ht!]
\centering
\begin{tabular}{ccc}
\includegraphics[width=0.32\textwidth]{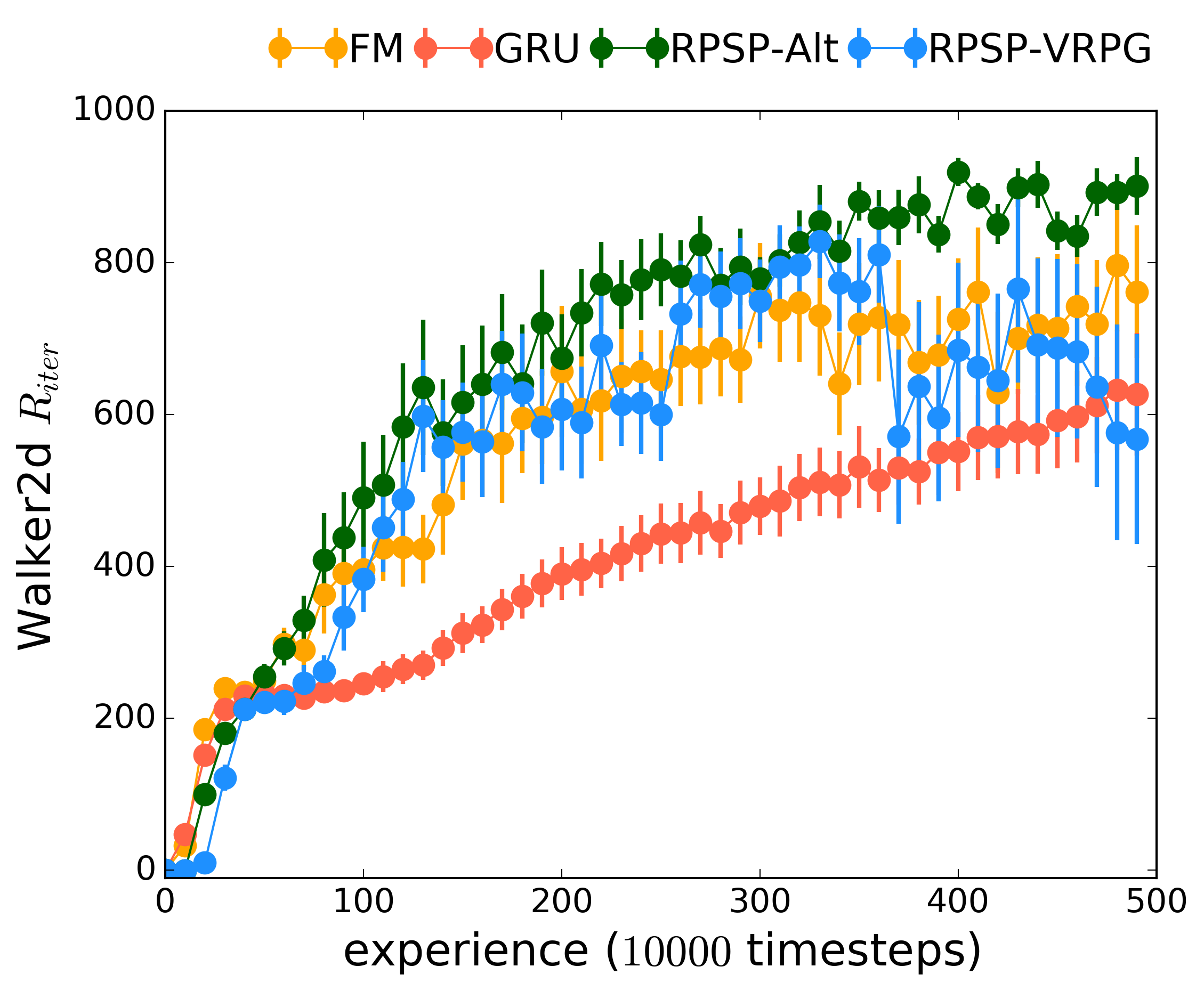} &  
\includegraphics[width=0.32\textwidth]{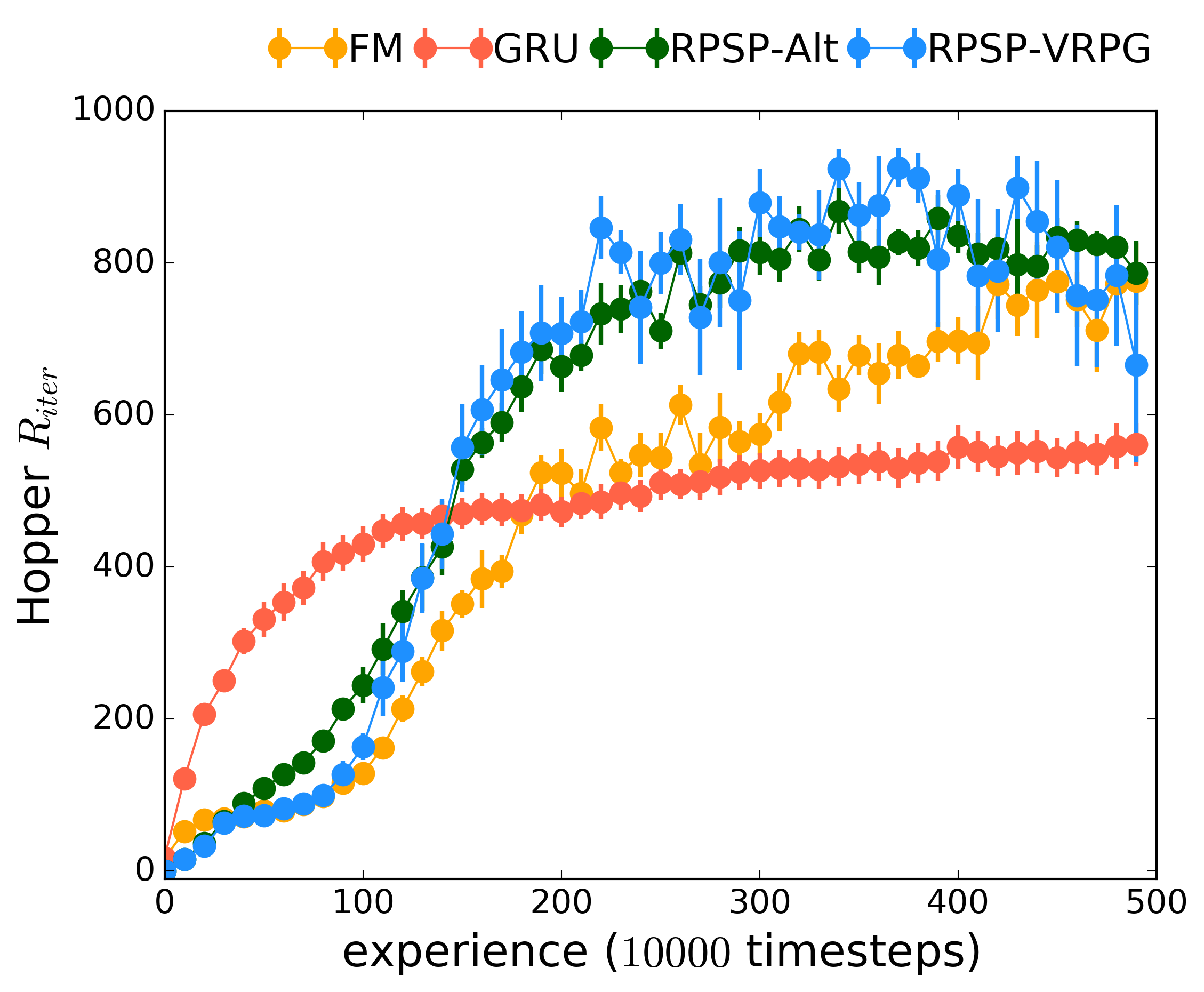} & 
\includegraphics[width=0.32\textwidth]{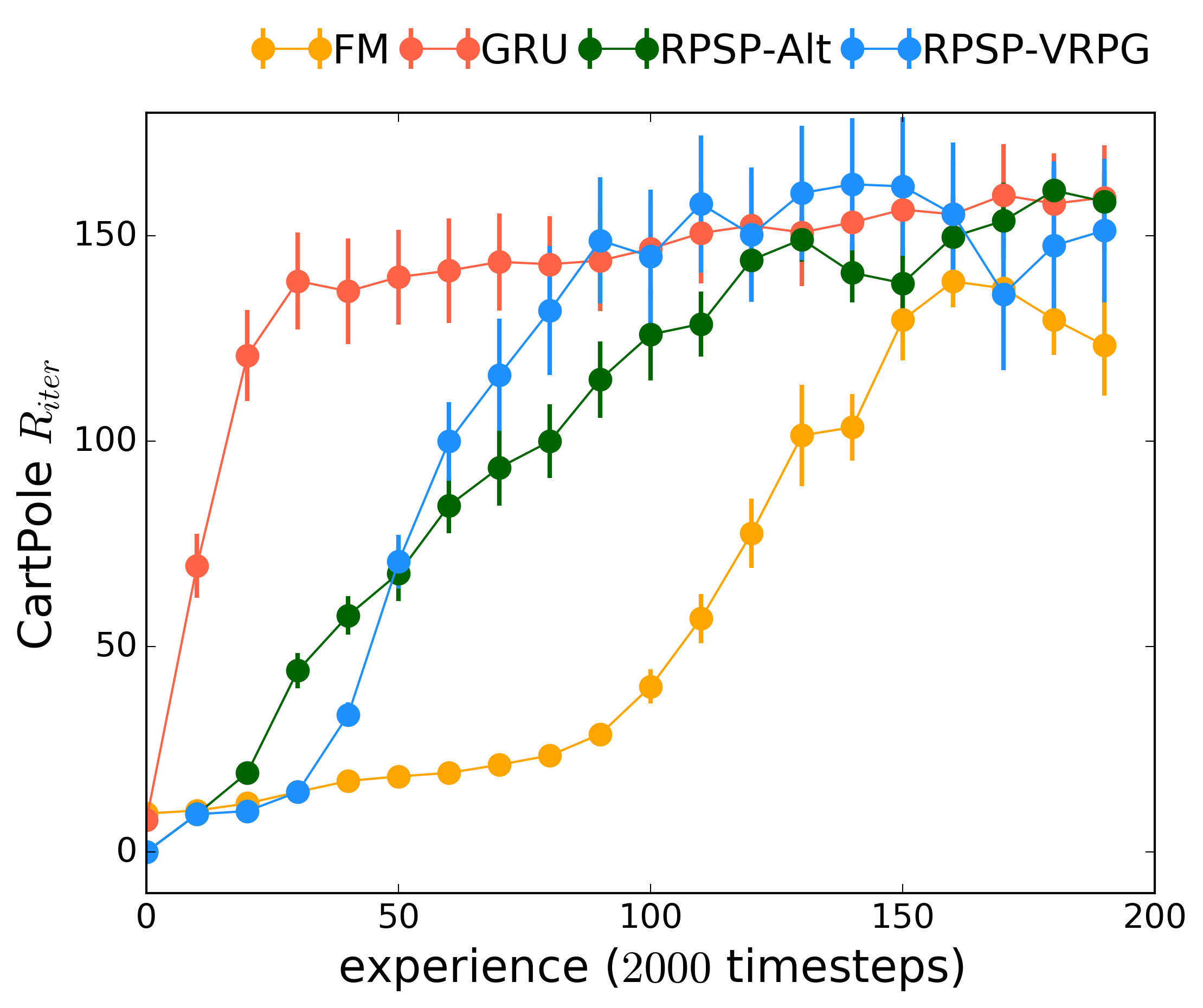} \\
(a) & (b) & (c) \\
\includegraphics[width=0.32\textwidth]{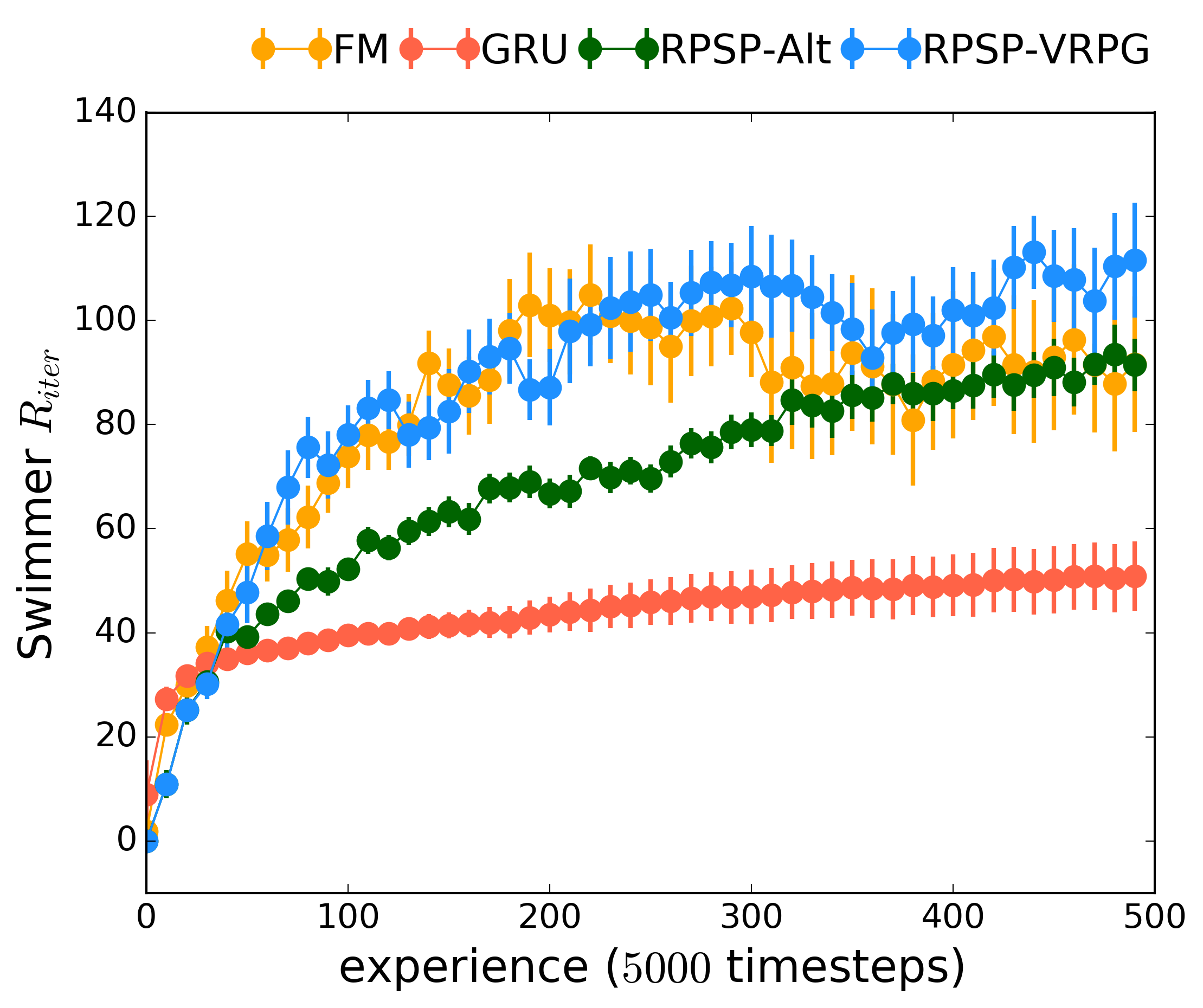} &
\includegraphics[width=0.32\textwidth, trim={0.2cm 0 0.1cm 0},clip]{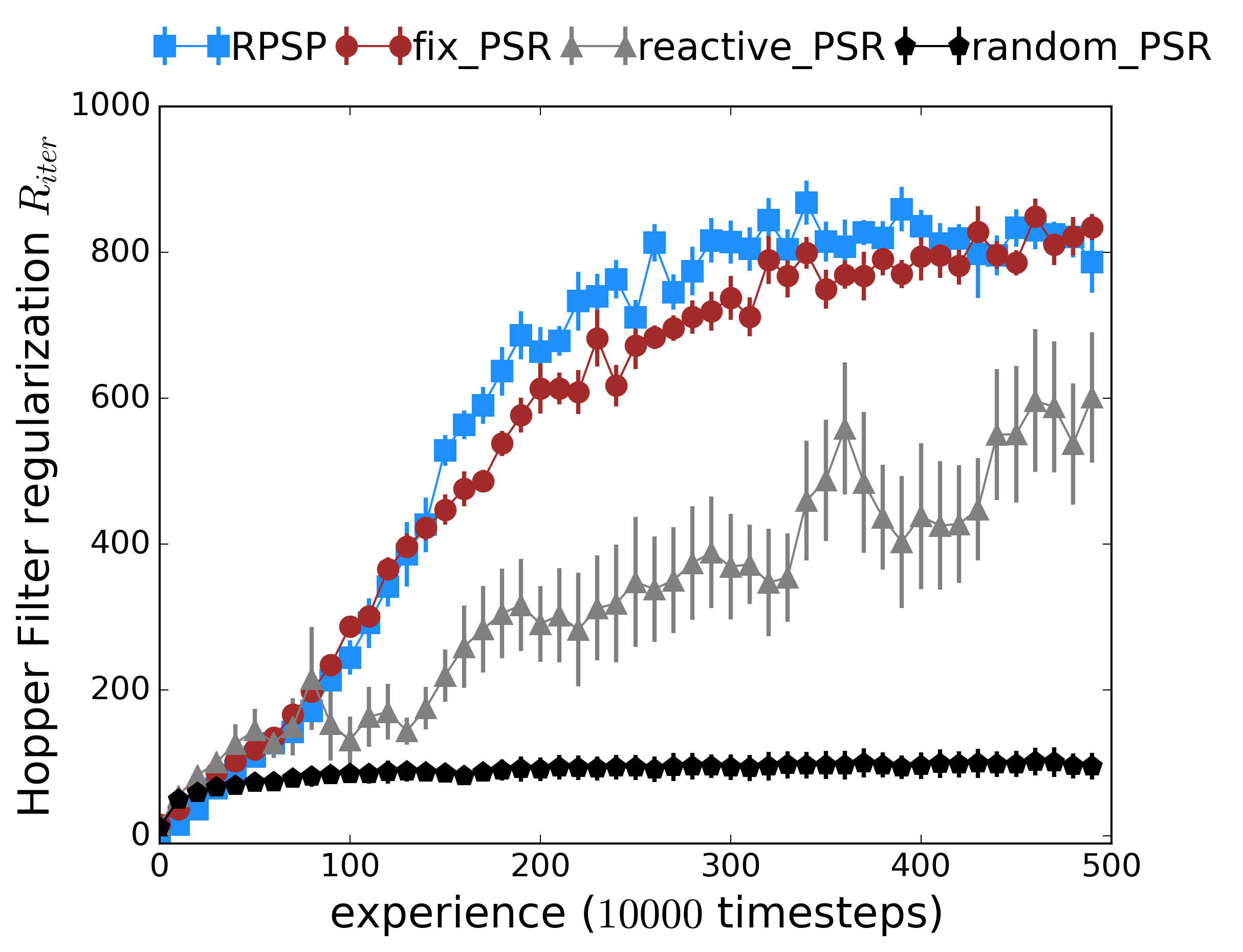} &
\includegraphics[width=0.32\textwidth, trim={0.1cm 0 0.2cm 0},clip]{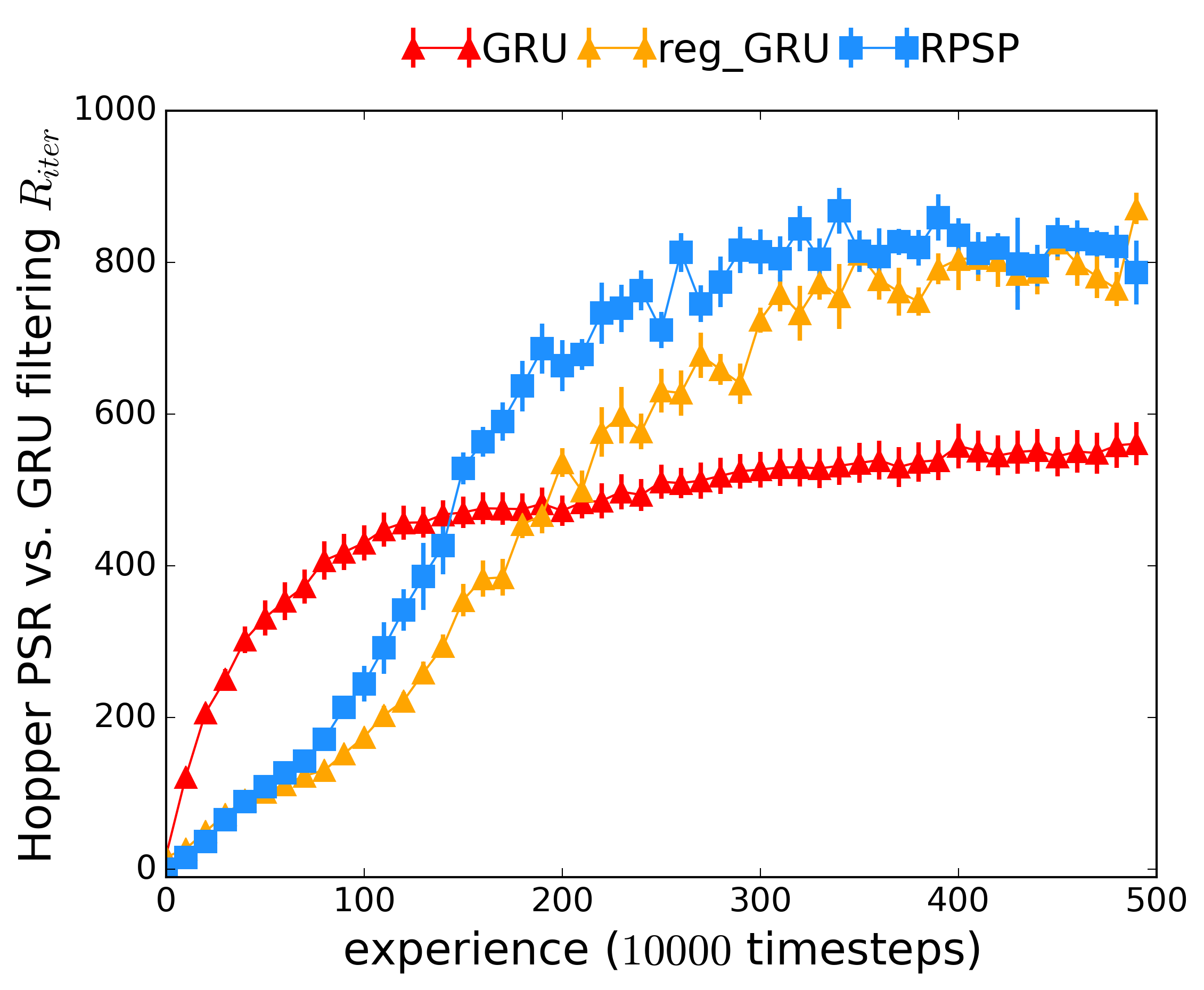} \\
(d) & (e) & (f) \\
\includegraphics[width=0.32\textwidth, trim={0.1cm 0 0.2cm 0},clip]{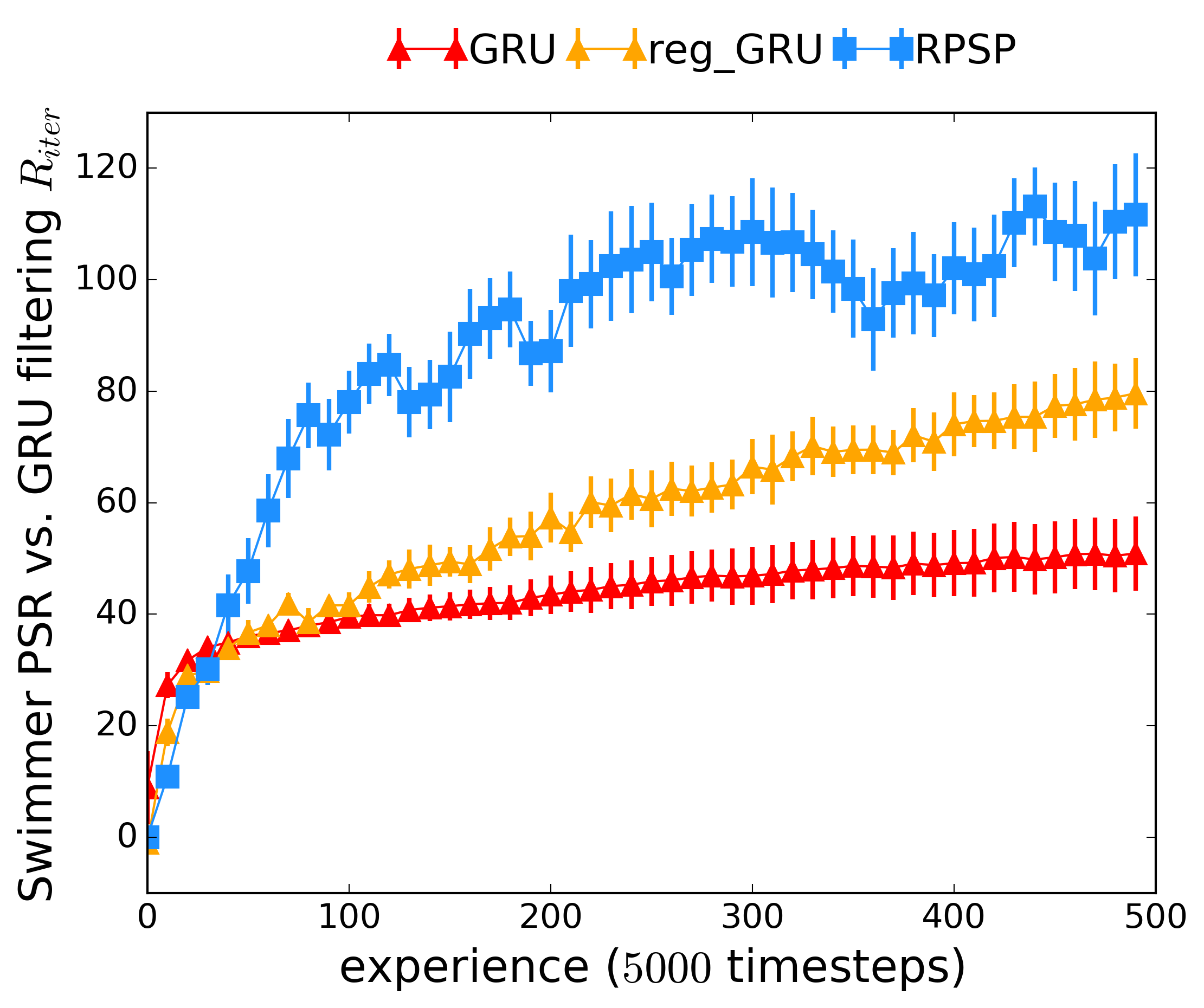} &
\multicolumn{2}{c}{
\includegraphics{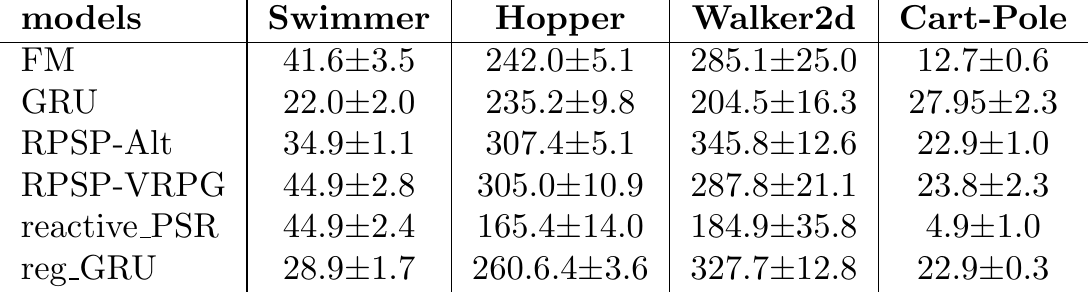}}\\
(g) & \multicolumn{2}{c}{(h)}
\end{tabular}
\caption{
Empirical average return over 10 epochs (bars indicate standard error). \textbf{(a-d):} Finite memory model $w=2$ (FM), GRUs (GRU), best performing RPSP with joint optimization (RPSP-VRPG) and best performing RPSP with alternate optimization (RPSP-Alt) on four environments.
\textbf{(e):} RPSP variations: fixed PSR parameters (fix\_PSR), without prediction regularization (reactive\_PSR), random initialization (random\_PSR).
\textbf{(f-g):} Comparison with GRU + prediction regularization (reg\_GRU).
RPSP graphs are shifted to the right to reflect the use of extra trajectories 
for initialization. \textbf{(h)}: Cumulative rewards (area under curve).}
\label{fig:rpsp}
\end{figure*}

\section{Experiments}
\label{sec:experiments}
We evaluate the RPSP-network's performance on a collection of reinforcement learning tasks using OpenAI Gym Mujoco environments.
\footnote{
\url{https://gym.openai.com/envs\#mujoco}}
We use default environment settings.
We consider only partially observable environments: only the angles of the joints of the agent are visible to the network, without velocities.

\label{sec:models}

\textbf{Proposed Models:}
We consider an RPSP with a predictive component based on RFFPSR, as described in \S\ref{sec:psr}
and \S\ref{sec:policynetwork}.
For the RFFPSR, we use 1000 random Fourier features on observation and action sequences followed by a PCA dimensionality reduction step to $d$ dimensions. 
We report the results for the best choice of $d \in \{10,20,30\}$.

We initialize the RPSP with two stage regression on a batch of $M_i$ initial trajectories (100 for Hopper, Walker and Cart-Pole, and 50 for Swimmer)
(equivalent to 10 extra iterations, or 5 for Swimmer).
We then experiment with both joint VRPG optimization (\textbf{RPSP-VRPG}) described in \S\ref{sec:joint}
and alternating optimization (\textbf{RPSP-Alt}) in \S\ref{sec:alt}. 
For RPSP-VRPG, we use the gradient normalization described in \S\ref{sec:normalization}.

Additionally, we consider an extended variation (\textbf{+obs}) that concatenates the predictive state with a window $w$ of previous observations as an extended form of predictive state $\tilde{\vec{q}}_t=[\vec{q}_t,\vec{o}_{t-w:t}]$. If PSR learning succeeded perfectly, this extra information would be unnecessary; however we observe in practice that including observations help the model learn faster and more stably. Later in the results section we report the RPSP variant that performs best. We provide a detailed comparison of all models in the appendix.

\textbf{Competing Models:}
\label{sec:baselines}
We compare our models to a finite memory model (\textbf{FM})
and gated recurrent units (\textbf{GRU}).
The finite memory models are analogous to RPSP, but replace the predictive state with a window of past observations. We tried three variants, FM1, FM2 and FM5, with window size of 1, 2 and 5 respectively (FM1 ignores that the environment is partially observable).
We compare to GRUs with $16,\ 32,\ 64$ and $128$-dimensional hidden states. We optimize network parameters using the \emph{RLLab}\footnote{\url{https://github.com/openai/rllab}}
implementation of TRPO with two different learning rates ($\eta=10^{-2},\ 10^{-3}$).

In each model, we use a linear baseline for variance reduction where the state of
the model (i.e. past observation window for FM, latent state for GRU and predictive state for RPSP) is used as the predictor variable.

\begin{figure*}[ht]
\centering
\includegraphics[trim={0 1cm 0 0},scale=0.3]{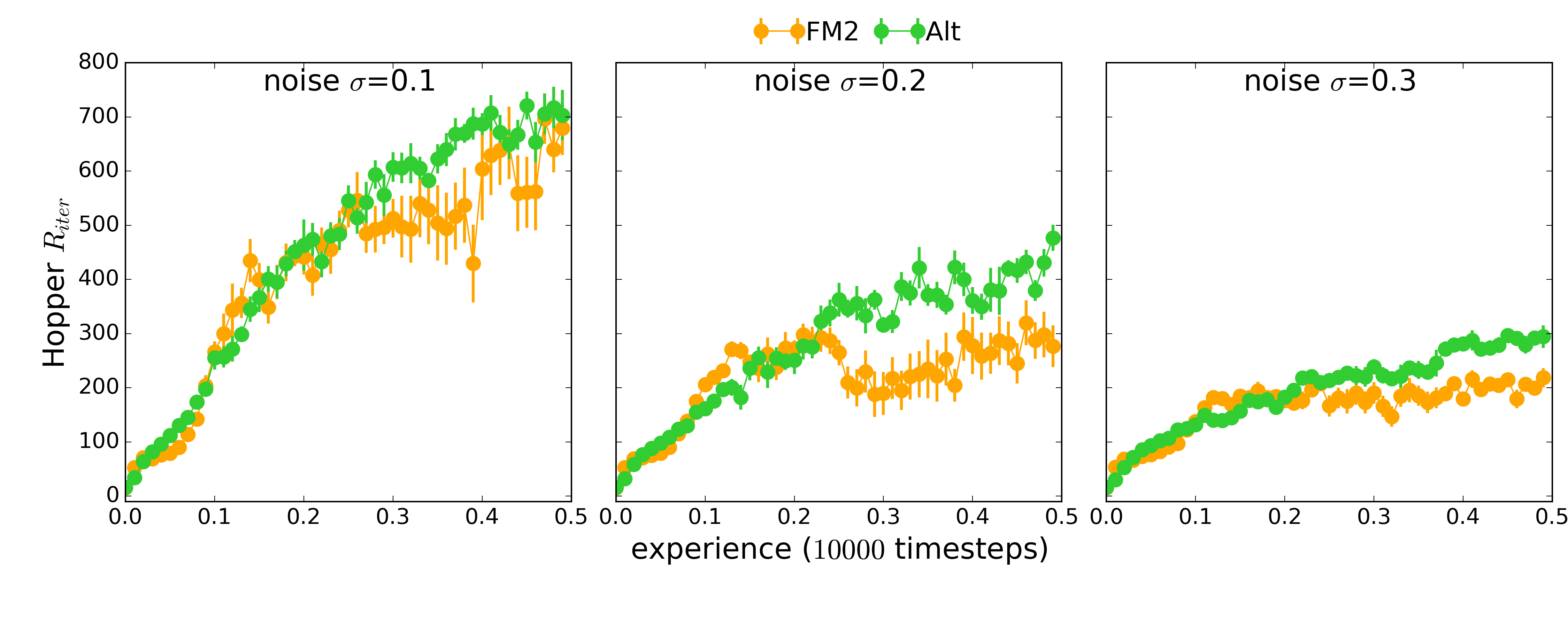}
\caption{Empirical average return over 10 trials with a batch of $M=10$ trajectories of $T=1000$ time steps for Hopper. (Left to right) Robustness to observation Gaussian noise $\sigma=\{0.1,\ 0.2,\ 0.3\}$, best RPSP with alternate loss (Alt) and Finite Memory model (FM2).}
\label{fig:noise}
\end{figure*}

\textbf{Evaluation Setup:}
\label{sec:metrics}
We run each algorithm for a number of iterations based on the environment (see \figref{fig:rpsp}).
After each iteration, we compute the average return $R_{iter}=\frac{1}{M} \sum_{m=1}^{M} \sum_{j=1}^{T_m} r_m^j$ on a batch of $M$ trajectories,
where $T_m$ is the length of the $m^{th}$ trajectory.
We repeat this process using 10 different random seeds and report the 
average and standard deviation of $R_{iter}$ over the 10 seeds for each iteration. 

For each environment, we set the number of samples in the batch to 10000 and the maximum length of each episode to 200, 500, 1000, 1000 for Cart-Pole, Swimmer, Hopper and Walker2d respectively.\footnote{For example, for a 1000 length environment we use a batch of 10 trajectories resulting in 10000 samples in the batch.} 

For RPSP, we found that a step size of $10^{-2}$ performs well for both VRPG and alternating 
optimization in all environments.
The reactive policy contains one hidden layer of 16 nodes with ReLU activation.
For all models, we report the results for the choice of hyper-parameters
that resulted in the highest mean cumulative reward (area under curve).

\section{Results and Discussion}
\label{sec:results}
\textbf{Performance over iterations:}
\figref{fig:rpsp} shows the empirical average return vs. the amount of interaction 
with the environment (experience), measured in time steps. 
For RPSP networks, the plots are shifted to account for the initial trajectories used to initialize the RPSP filtering layer. The amount of shifting is equivalent to 
10 trajectories.

We observe that RPSP networks (especially RPSP-Alt) perform well in every environment,
competing with or outperforming the top model in terms of the learning speed and the final reward, with the exception of Cart-Pole where the gap to GRU is larger. We report the cumulative reward for all environments in Table~\ref{fig:rpsp}(h). For all except Cart-Pole, come variant of RPSP is the best performing model. For Swimmer our best performing model is only statistically better than FM model (t-test, $p<0.01$), while for Hopper our best RPSP model performs statistically better than FM and GRU models (t-test, $p<0.01$) and for Walker2d RPSP outperforms only GRU baselines (t-test, $p<0.01$). For Cart-Pole the top RPSP model performs better than the FM model (t-test, $p<0.01$) and it is not statistically significantly different than the GRU model.
We also note that RPSP-Alt provides similar performance to the joint optimization (\textbf{RPSP-VRPG}), but converges faster.

\textbf{Effect of proposed contributions:}
Our RPSP model is based on a number of components:
(1) State tracking using PSR (2) Consistent initialization using two-stage regression (3) End-to-end training of state tracker and policy 
(4) Using observation prediction loss to regularize training.

We conducted a set of experiments to verify the benefit of each component.\footnote{
Due to space limitation, we report the results of these experiments on Hopper environment. Results for other environments can be found in the supplementary material.
}
In the first experiment, we test three variants of RPSP: one where the PSR is randomly initialized (\textbf{random\_PSR}), another one where the PSR is fixed at the initial value (\textbf{fix\_PSR}),
and a third one where we train the RPSP network without prediction loss regularization (i.e. we set $\alpha_2$ in \eqref{eq:losspsr}) to 0 (\textbf{reactive\_PSR}).
\figref{fig:rpsp}(e) demonstrates that these variants are inferior to our model, showing the importance of two-stage initialization, end-to-end training and observation prediction loss respectively.

In the second experiment, we replace the PSR with a GRU that is initialized using backpropagation through time. This is analogous to the predictive state decoders proposed in \citep{psd}, where observation prediction loss is included when optimizing a GRU policy network (\textbf{reg\_GRU}).\footnote{
The results we report here are for the partially observable
setting which is different from the reinforcement learning experiments in \citep{psd}.}
\figref{fig:rpsp}(f-g) shows that a GRU model is inferior to a PSR model, where the initialization procedure is consistent and does not suffer from local optima.


\textbf{Effect of observation noise:}
We also investigated the effect of observation noise on the RPSP model and the 
competitive FM baseline by applying Gaussian noise of increasing variance to
observations.

\figref{fig:noise} shows that
while FM was very competitive with RPSP in the noiseless case,
RPSP has a clear advantage over FM in the case of mild noise.
However, the performance gap vanishes if excessive noise is applied.

\section{Conclusion}
\label{sec:discussion}
We propose RPSP-networks, combining ideas from predictive state representations and recurrent networks for reinforcement learning. We use PSR learning algorithms to provide a statistically consistent initialization of the state tracking component, and propose gradient-based methods to maximize 
expected return while reducing prediction error.

We compare RPSP against different baselines and empirically show the efficacy of the proposed approach in terms of speed of convergence and overall expected return.


\bibliographystyle{icml2018}
\bibliography{psr_rl}

\clearpage
\section*{Appendix}

\section{Learning Predictive States with two stage regression}
\label{sec:ivr}
In this section we describe the process of initializing the state tracking part of RPSP-networks. We derive the state filter equation for RFFPSR representation using 2-stage regression, for a complete derivation refer to \citet{hefny:17}.  
Let $\vec{a}_{1:t-1}=\vec{a}_1,\dots,\vec{a}_{t-1}\in \mathcal{A}^{t-1}$ be the set of actions performed by an agent, followed by observations $\vec{o}_{1:t-1}=\vec{o}_1,\dots,\vec{o}_{t-1}\in \mathcal{O}^{t-1}$ received by the environment up to time $t$. Together they compose the entire \emph{history} up to time $t$ $\vec{h}^{\infty}_t \equiv \{\vec{a}_{1:t-1}, \vec{o}_{1:t-1}\}$. Consider \emph{future} to be a sequence of consecutive $k$-observations $\vec{o}_{t:t+k-1}\in \mathcal{O}^{k}$, and let us define the  feature mappings shown in \figref{fig:windows}, for immediate, future and extended future actions and observations:
\begin{itemize}[leftmargin=.2in, itemsep=-1pt,topsep=1pt, partopsep=0pt]
\item[--]$\phi^o_t(\vec{o}_t):\mathcal{O}\rightarrow \mathbb{R}^{O}$ of immediate observations,
\item[--]$\phi^a_t(\vec{a}_t):\mathcal{A}\rightarrow \mathbb{R}^{A}$ of immediate actions,
\item[--]$\bs{\psi}^o_t(\vec{o}_{t:t+k-1}):\mathcal{O}^k\rightarrow \mathbb{R}^{F^O}$ of future observations,
\item[--]$\psi^a_t(\vec{a}_{t:t+k-1}):\mathcal{A}^k\rightarrow \mathbb{R}^{F^A}$ of future actions, 
\item[--]$\xi^o_t(\vec{o}_{t:t+k})\equiv [\bs{\psi}^o_t \otimes \bs{\phi}^o_t,\bs{\phi}^o_t\otimes\bs{\phi}^o_t]  :\mathcal{O}^{k+1}\rightarrow \mathbb{R}^{(F^O+O}\otimes \mathbb{R}^{2O}$ of extended observations,
\item[--]$\xi^a_t(\vec{a}_{t:t+k})\equiv \bs\psi^a_t \otimes \bs{\phi}^a_t :\mathcal{A}^{k+1}\rightarrow \mathbb{R}^{F^A}\otimes \mathbb{R}^{A}$ of extended actions.
\end{itemize}

We assume future expectations $\bar{\vec{p}}_t=\mathbb{E}[\bs{\xi}^o_t|\mathbf{do}(\vec{a}_{t:t+k})]$ are a linear transformation of extended future expectations $\bar{\vec{q}}_t=\mathbb{E}[\bs{\psi}^o_t|\bold{do}(\vec{a}_{t:t+k-1})]$, in~\figref{fig:windows}. 
Due to temporal correlation between these two expectations $\bar{\vec{p}}_t$ and $\bar{\vec{q}}_t$, we cannot directly learn the mapping $W_\mathrm{ext}$ over empirical estimates of $\bar{\vec{p}}_t$ and $\bar{\vec{q}}_t$, using linear regression, since their noise is also correlated.
Alternatively, we turn to an instrumental variable regression where history is used as instrument, since it is correlated with the observables $\bs{\psi}_t$ and $\bs{\xi}_t$, but not with the noise. We go from predictive to extended states by first computing a possibly non-linear regression from histories, to both predictive \eref{eq:Stage1a} and extended statistics \eref{eq:Stage1b}  (stage-1 regression $a$ and $b$). In a Hilbert Space Embedding PSR (HSE-PSR) this non-linear regression can be computed by Kernel Bayes Rule (KBR)~\citep{fukumizu13}. 
\begin{align}
\vec{p}_t&\equiv\mathbb{E}\left[\bs{\xi}^o_{t+1}|\bs{\xi}^a_{t+1};\vec{h}_{t}^{\infty}\right] \ \text{stage-1$a$}\label{eq:Stage1a}\\ 
\vec{q}_t&\equiv\mathbb{E}\left[\bs{\psi}^o_t |\bs{\psi}^a_t;\vec{h}_{t}^{\infty}\right] \ \ \ \ \ \ \ \text{stage-1$b$}\label{eq:Stage1b}
\end{align}
Subsequently, we linearly regress the denoised extended state $\vec{p}_t$ from the denoised predictive state $q_t$, using a least squares approach (stage-2 regression), in \eref{eq:prediction}. For ease of explanation let us further partition extended states in two parts $\vec{p}_t\equiv[\vec{p}_t^\xi, \vec{p}_t^o]$ and $W_\mathrm{ext}\equiv[W_\mathrm{ext}^\xi, W_\mathrm{ext}^o]$, derived from the skipped future $\bs{\psi}_{t+1}^o$ and present $\bs{\phi}^o_t$ observations, see \figref{fig:windows}. 
\begin{align}\notag
\vec{p}^{\xi}_t&\equiv\mathbb{E}\left[\bs{\psi}^o_{t+1}\otimes \bs{\phi}^o_t|\bs{\psi}^a_{t+1}\otimes \bs{\phi}^a_t,\vec{h}_{t}^{\infty}\right]\\ \notag
\vec{p}^o_t&\equiv\mathbb{E}\left[\bs{\phi}^o_t \otimes \bs{\phi}^o_t |\bs{\phi}^a_t,\vec{h}_{t}^{\infty}\right]\\
\vec{p}_t&= [\vec{p}_t^{\xi}, \vec{p}_t^o]^\top=[W^{\xi}_{sys}, W_\mathrm{ext}^o]^\top \vec{q}_t\ \ \text{stage-2}\label{eq:prediction}
\end{align}
Stage-1 regression, for RFFPSRs can be derived from either a joint regression over action/observation pairs, using KBR or by solving a regularized least squares problem, for a full derivation refer to \citet{hefny:17}.\\
The second step in \eref{eq:filter_cond} is provided by a \emph{filtering function} $f_{t}$, that tracks predictive states over time:
\begin{eqnarray}
\vec{q}_{t+1}= f_\mathrm{cond} (W_\mathrm{ext}\vec{q}_t, \vec{o}_t, \vec{a}_t) \equiv f_{t} (\vec{q}_t, \vec{o}_t, \vec{a}_t)\label{eq:filtering}
\end{eqnarray}
Namely, for HSE-PSRs filtering corresponds to conditioning on the current observation via KBR, in \eref{eq:KBR}. Here, we define $\vec{p}_t^\xi$ as a 4-mode tensor $(\bs{\psi}^o_{t+1}, \bs{\phi}^o_t, \bs{\psi}^a_{t+1},\bs{\phi}^a_t)$, where $\times_{\phi^o}$ defines the multiplication along the ${\phi}^o$-mode.
\begin{align}
\vec{q}_{t+1} &= \mathbb{E}[\bs{\psi}^o_{t+1}|\bs{\psi}^a_{t+1};\vec{h}_{t+1}^{\infty}] \label{eq:KBR}\\ \notag
&= \mathbb{E}[\bs{\psi}^o_{t+1}|\bold{do}(\bs{\psi}^a_{t+1}, \bs{\phi}^a_t),\bs{\phi}^o_t; \vec{h}_{t}^{\infty}] \\ \notag
&= \mathbb{E}[\bs{\psi}^o_{t+1}\otimes \bs{\phi}^o_t|\bs{\psi}^a_{t+1}, \bs{\phi}^a_t;\vec{h}_{t}^{\infty}] \times_{\bs{\phi}^o} \bs{\phi}^o_t \times_{\bs{\phi}^a} \bs{\phi}^a_t \\\notag
&\tab[2cm] \left[\mathbb{E}[\bs{\phi}^o_t \otimes \bs{\phi}^o_t |\bs{\phi}^a_t;\vec{h}_{t}^{\infty}] {\bs{\phi}^a_t}^\top  +\lambda I\right]^{-1} \\ \notag
&= \vec{p}^{\xi}_t \times_{\bs{\phi}^o} \bs{\phi}^o_t \times_{\bs{\phi}^a} \bs{\phi}^a_t \left[ \vec{p}_t^o {\bs{\phi}^a_t}^\top  +\lambda I \right]^{-1}\\ \notag
&=W_\mathrm{ext}^\xi \vec{q}_t \times_{\bs{\phi}^o} \bs{\phi}^o_t \times_{\bs{\phi}^a} \bs{\phi}^a_t \left[ W_\mathrm{ext}^o \vec{q}_t {\bs{\phi}^a_t}^\top  +\lambda I \right]^{-1} \\ \notag
&\equiv f_t( \vec{q}_t, \vec{a}_t,\vec{o}_t)
\end{align}
This equation explicitly defines the full predictive state update equation or filtering in \eref{eq:filter_cond}, when considering HSE-PSR's representation of $\vec{p}_t$ and $\vec{q}_t$.

\section{RPSP-network optimization algorithms}
\label{sec:updatealgs}
For clarity we provide the pseudo-code for the joint and alternating update steps defined in the $UPDATE$ step in \aref{alg:psrnetwork}, in section \S\ref{sec:policygradient}. We show the joint VRPG update step in \aref{alg:vrpg}, and the alternating (Alternating Optimization) update in \aref{alg:altop}.

\begin{algorithm}[t]
   \caption{UPDATE (VRPG)}
   \label{alg:vrpg}
\small
\begin{algorithmic}[1]
\STATE \textbf{Input:} $\boldsymbol{\theta}^{n-1}$, trajectories $\mathcal{D}$=$\{\tau^i\}_{i=1}^M$, and learning rate $\eta$.
\STATE Estimate a linear baseline $b_t=\vec{w}_{b}^\top \vec{q}_t$, from the expected reward-to-go function for the batch $\mathcal{D}$: \par
$\tab\vec{w}_{b}=\mathrm{arg}\min\limits_{\vec{w}} \left \|\frac{1}{TM}\sum\limits_{i=1}^{M} \sum\limits_{t=1}^{T_i} R_t(\tau^i_t) - \vec{w}^\top\vec{q}_t\right\|^2$.

\STATE Compute the VRPG loss gradient w.r.t. $\boldsymbol{\theta}$, in \eref{eq:reinf}: \par
$\ \ \nabla_{\boldsymbol{\theta}}\ell_1=  \frac{1}{M} \sum\limits_{i=1}^M\sum\limits_{t=0}^{T_i}\nabla_{\boldsymbol{\theta}}\log \pi_{\boldsymbol{\theta}}(\vec{a}_t^i|\vec{q}_t^i) (R_t(\tau^i)-b_t)$.

\STATE Compute the prediction loss gradient: \par
$\ \ \nabla_{\boldsymbol{\theta}}\ell_2=\frac{1}{M}\sum\limits_{i=1}^M\sum\limits_{t=1}^{T_i}
\nabla_{\boldsymbol{\theta}} \left\| W_{pred} (\vec{q}_t^i \otimes \vec{a}_t^i)- \vec{o}_{t}^i \right\|^2$.
\STATE Normalize gradients $\nabla_{\boldsymbol{\theta}} \ell_j=\textsc{NORMALIZE}(\boldsymbol{\theta},\ell_j)$, in \eref{eq:normalization}.
\STATE Compute joint loss gradient as in \eref{eq:losspsr}: \par
$\tab\nabla_{\boldsymbol{\theta}}\mathcal{L}= \alpha_1\nabla_{\boldsymbol{\theta}}\ell_1+\alpha_2\nabla_{\boldsymbol{\theta}}\ell_2$.
\STATE Update policy parameters: $\ \ \boldsymbol{\theta}^{n} =\ \textsc{ADAM}(\boldsymbol{\theta}^{n-1},\nabla_{\boldsymbol{\theta}}\mathcal{L}, \eta)$
\STATE \textbf{Output:} Return $\boldsymbol{\theta}^n=(\boldsymbol{\theta}^n_\mathrm{PSR}, \boldsymbol{\theta}^n_\mathrm{re}, \eta)$.
\end{algorithmic}
\end{algorithm}

\begin{algorithm}[t]
   \caption{UPDATE (Alternating Optimization)}
   \label{alg:altop}
\small
\begin{algorithmic}[1]
\STATE \textbf{Input:} $\boldsymbol{\theta}^{n-1}$, trajectories $\mathcal{D}=\{\tau^i\}_{i=1}^M$.
\STATE Estimate a linear baseline $b_t=\vec{w}_{b}^\top \vec{q}_t$, from the expected reward-to-go function for the batch $\mathcal{D}$: \par
$\tab\vec{w}_{b}=\mathrm{arg}\min\limits_{\vec{w}} \left \|\frac{1}{TM}\sum\limits_{i=1}^{M}\sum\limits_{t=1}^{T_i} R_t(\tau^i_t) - \vec{w}^\top\vec{q}_t\right\|^2$.
\STATE Update $\boldsymbol{\theta}_{\mathrm{PSR}}$ using the joint VRPG loss gradient in \eref{eq:losspsr}: \par
$\tab\boldsymbol{\theta}^n_{\mathrm{PSR}}\leftarrow\textsc{UPDATE VRPG}(\boldsymbol{\theta}^{n-1},\mathcal{D})$.
\STATE Compute descent direction for TRPO update of $\boldsymbol{\theta}_\mathrm{re}$: \par
$v = H^{-1} g$, where \par
$H = \nabla^2_{\boldsymbol{\theta}_\mathrm{re}} 
\sum\limits_{i=1}^M D_{KL}\left(\pi_{\boldsymbol{\theta}^{n-1}} (\vec{a}^i_t|\vec{q}^i_t)\mid\pi_{\boldsymbol{\theta}}(\vec{a}^i_t|\vec{q}^i_t)\right)
$, \par
$g = \nabla_{\boldsymbol{\theta}_\mathrm{re}} \dfrac{1}{M} \sum\limits_{i=1}^M\sum\limits_{t=1}^{T_i} \dfrac{\pi_{\boldsymbol{\theta}}(\vec{a}_t^i|\vec{q}_t^i)}{\pi_{\boldsymbol{\theta}^{n-1}}(\vec{a}_t^i|\vec{q}_t^i)} (R_t(\tau^i)-b_t)$.
\STATE Determine a step size $\eta$ through a line search on $v$ to maximize the objective in 
\eqref{eq:reactiveup} while maintaining the constraint.
\STATE $\boldsymbol{\theta}_{\mathrm{PSR}}^n \leftarrow \boldsymbol{\theta}_{\mathrm{PSR}}^{n-1} + \eta v$
\STATE \textbf{Output:} Return $\boldsymbol{\theta}^n=(\boldsymbol{\theta}^n_\mathrm{PSR}, \boldsymbol{\theta}^n_{\mathrm{re}})$.
\end{algorithmic}
\end{algorithm}

\section{Comparison to RNNs with LSTMs/GRUs}
\label{sec:rnnconnection}
RPSP-networks and RNNs both define recursive models that are able to retain information about previously observed inputs. 
BPTT for learning predictive states in PSRs bears many similarities with BPTT for training hidden states in LSTMs or GRUs. In both cases the state is updated via a series of alternate linear and non-linear transformations. For predictive states the linear transformation $\vec{p}_t=W_\mathrm{ext}\ \vec{q}_t$ represents the system prediction: from expectations over futures $\vec{q}_t$ to expectations over extended features $\vec{p}_t$. The non-linear transformation, in place of the usual activation functions (tanh, ReLU), is replaced by $f_\mathrm{cond}$ that conditions on the current action and observation to update the expectation of the future statistics $\vec{q}_{t+1}$ in \eref{eq:filter_cond}.
It is worth noting that these transformations represent non-linear state updates, as in RNNs, but where the form of the update is defined by the choice of representation of the state. For Hilbert Space embeddings it corresponds to conditioning using Kernel Bayes Rule, in \ref{eq:KBR}. An additional source of linearity is the representation itself. When we consider linear transformations $W_\mathrm{pred}$ and $W_\mathrm{ext}$ we refer to transformations between kernel representations, between Hilbert Space Embeddings.

PSRs also define computation graphs, where the parameters are optimized by leveraging the states of the system. Predictive states can leverage history like LSTMs/GRUs, PSRs also have memory, since they learn to track in the Reproducing Kernel Hilbert Space (RKHS) space of distributions of future observations based on past histories. PSRs provide the additional benefit of having a clear interpretation as a 
prediction of future observations and could be trained 
based on that interpretation. For this reason, RPSPs have a statistically driven form of initialization, that can be obtained using a moment matching technique, with good theoretical guarantees. In contrast RNNs may exploit heuristics for an improved initialization, such as , however, defining the best strategy does not guarantee good performance. In this paper the proposed initialization, provides guarantees in the infinite data assumption~\cite{hefny:17}.

\section{Additional Experiments}
\label{sec:exp2}
In this section, we investigate the effect of using different variants of RPSP networks, we test against a random initialization of the predictive layer, and provide further experimental evidence for baselines.

\begin{figure}[h]
\centering
\begin{subfigure}[t]{0.40\textwidth}
\includegraphics[width=\textwidth]{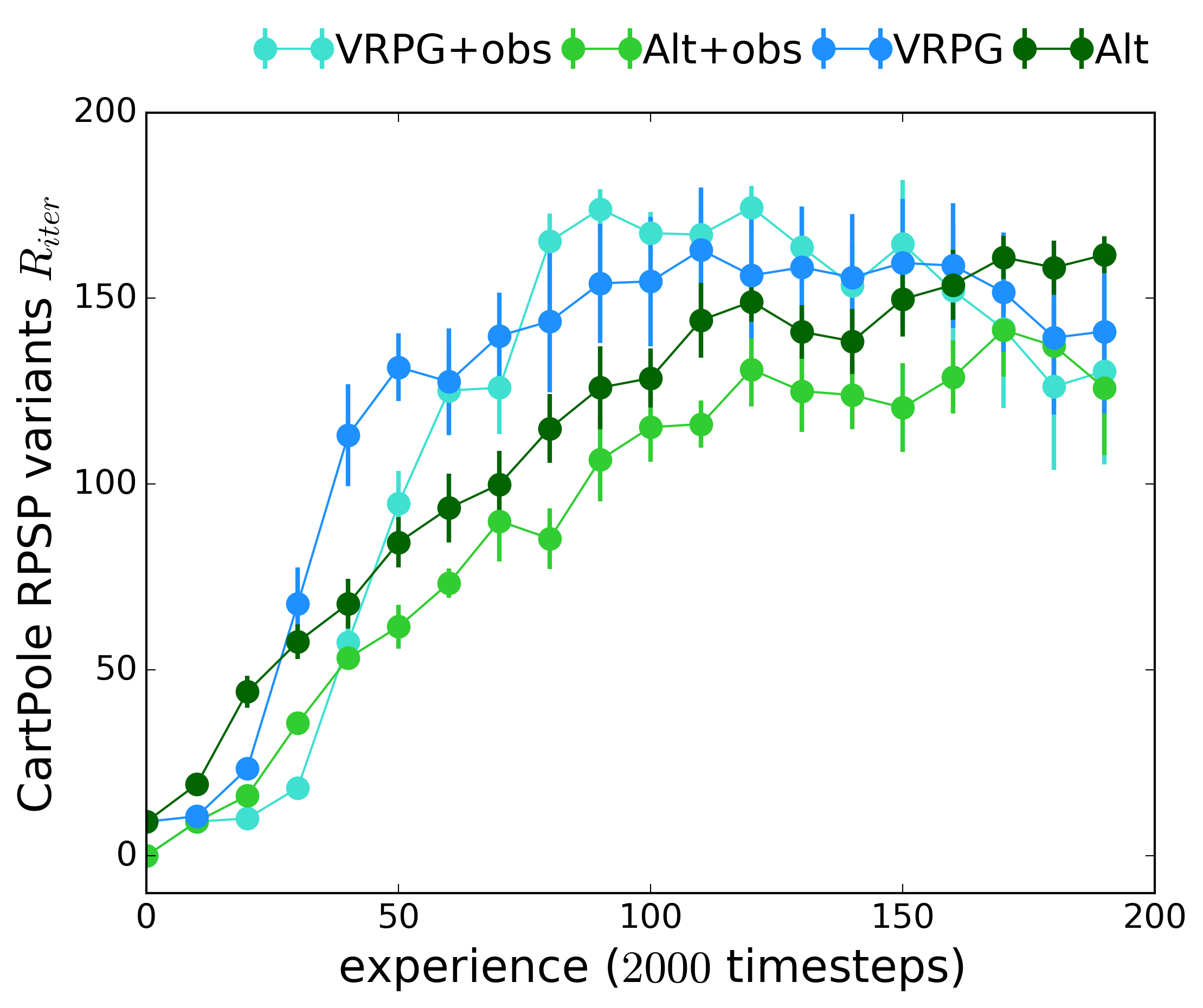}
\end{subfigure}
\begin{subfigure}[t]{0.40\textwidth}
\includegraphics[width=\textwidth]{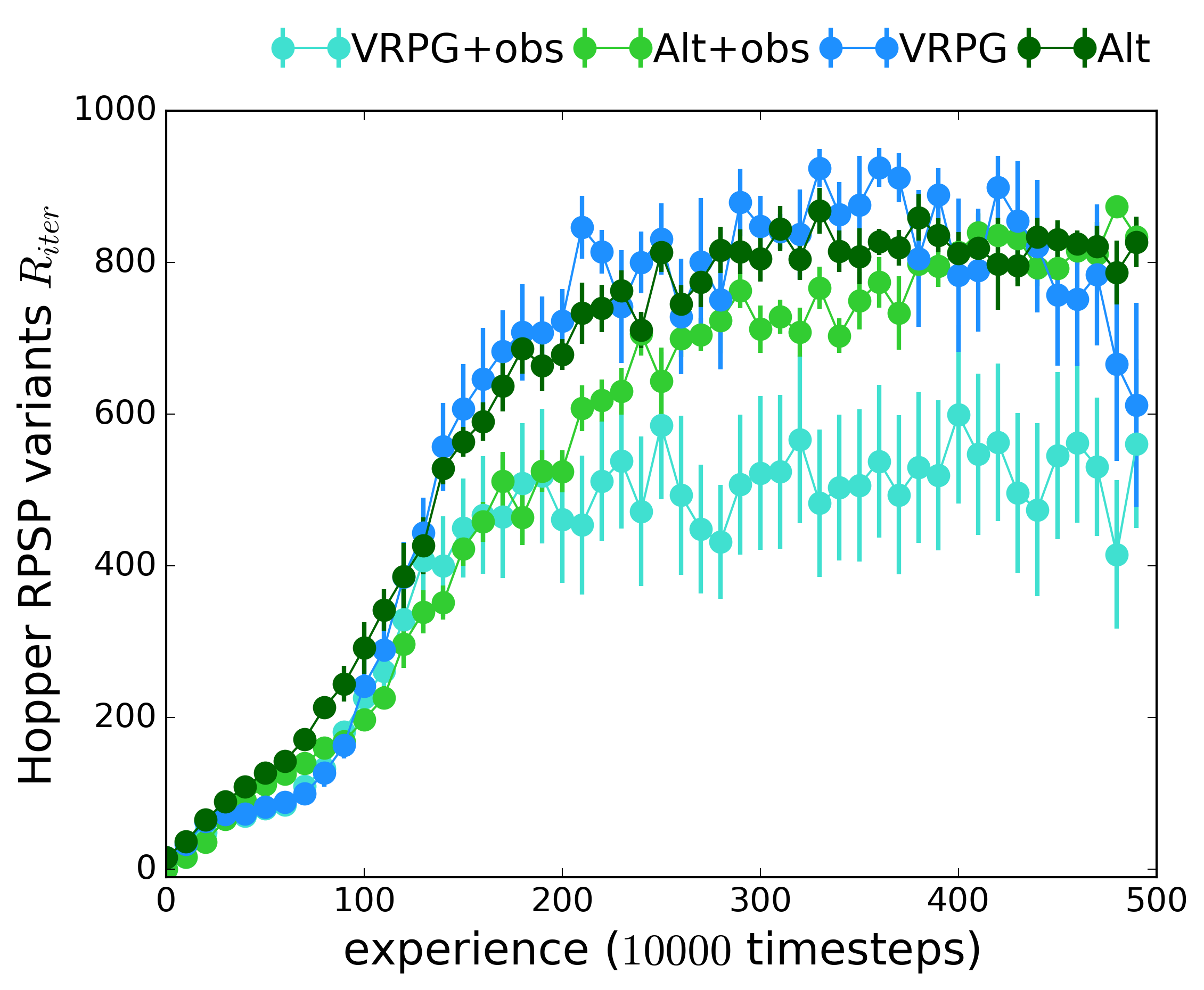}
\end{subfigure}
\begin{subfigure}[t]{0.40\textwidth}
\includegraphics[width=\textwidth]{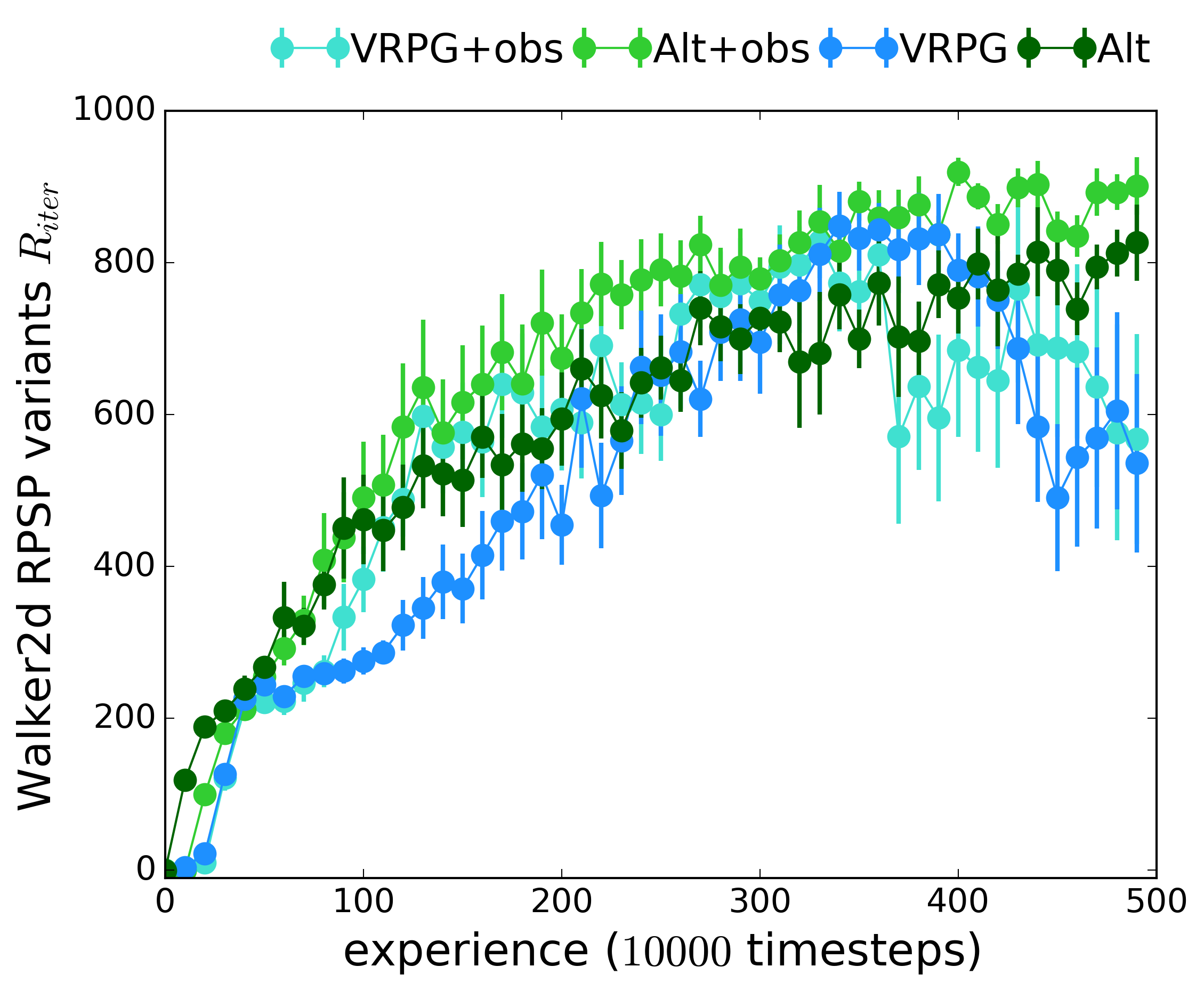}
\end{subfigure}
\begin{subfigure}[t]{0.40\textwidth}
\includegraphics[width=\textwidth]{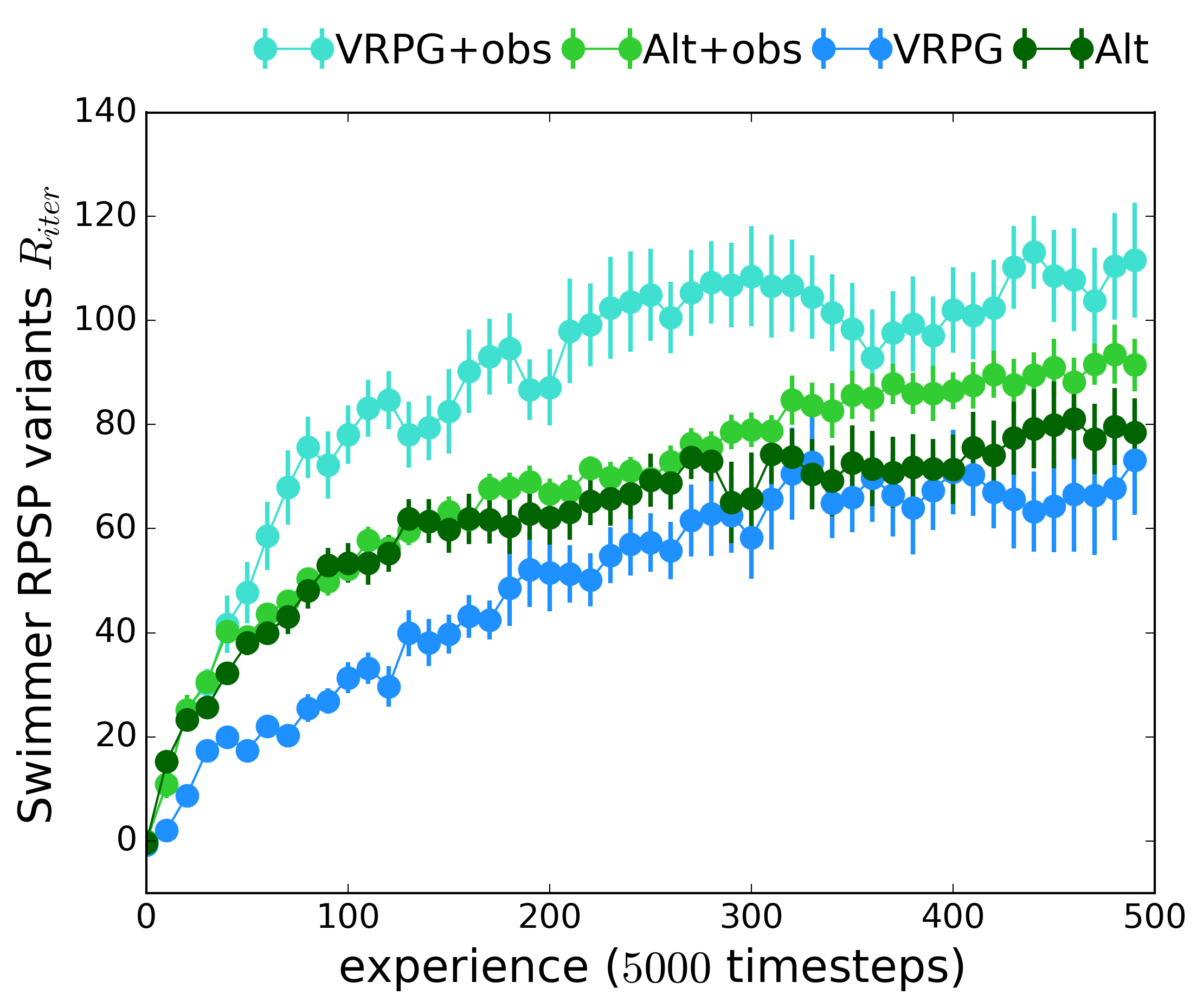}
\end{subfigure}
\caption{Reward over iterations for RPSP variants over a batch of $M=10$ trajectories and $10$ trials.}
\label{fig:rpspvar}
\end{figure}

\textbf{RPSP optimizers:}
Next, we compare several RPSP variants for all environments. We test the two RPSP variants, joint and alternate loss with predictive states and with augmented predictive states (\textbf{+obs}). The first variant is the standard ``vanilla'' RPSP, while the second variant is an RPSP with augmented state representation where we concatenate the previous window of observations to the predictive state (+obs). We provide a complete comparison of RPSP models using augmented states with observations for all environments in \figref{fig:rpspvar}. We compare with both joint optimization (VRPG+obs) and an alternating approach (Alt+obs). Extended predictive states with a window of observations ($w=2$) provide better results in particular for joint optimization. This extension might mitigate prediction errors, improving information carried over by the filtering states.

\begin{figure}[h!]
\centering
\begin{subfigure}[t]{0.4\textwidth}
\includegraphics[width=\textwidth]{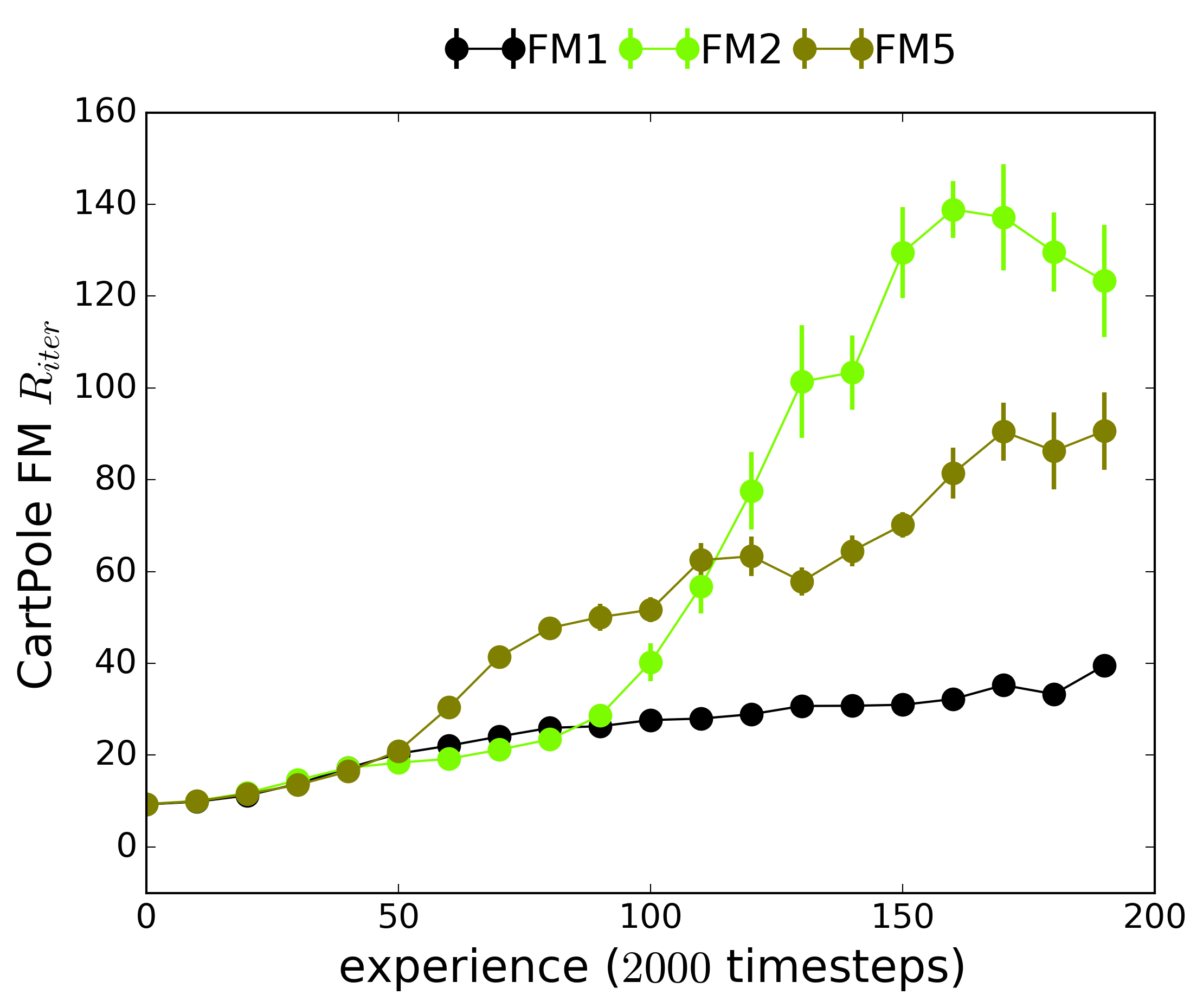}
\end{subfigure}
\begin{subfigure}[t]{0.4\textwidth}
\includegraphics[width=\textwidth]{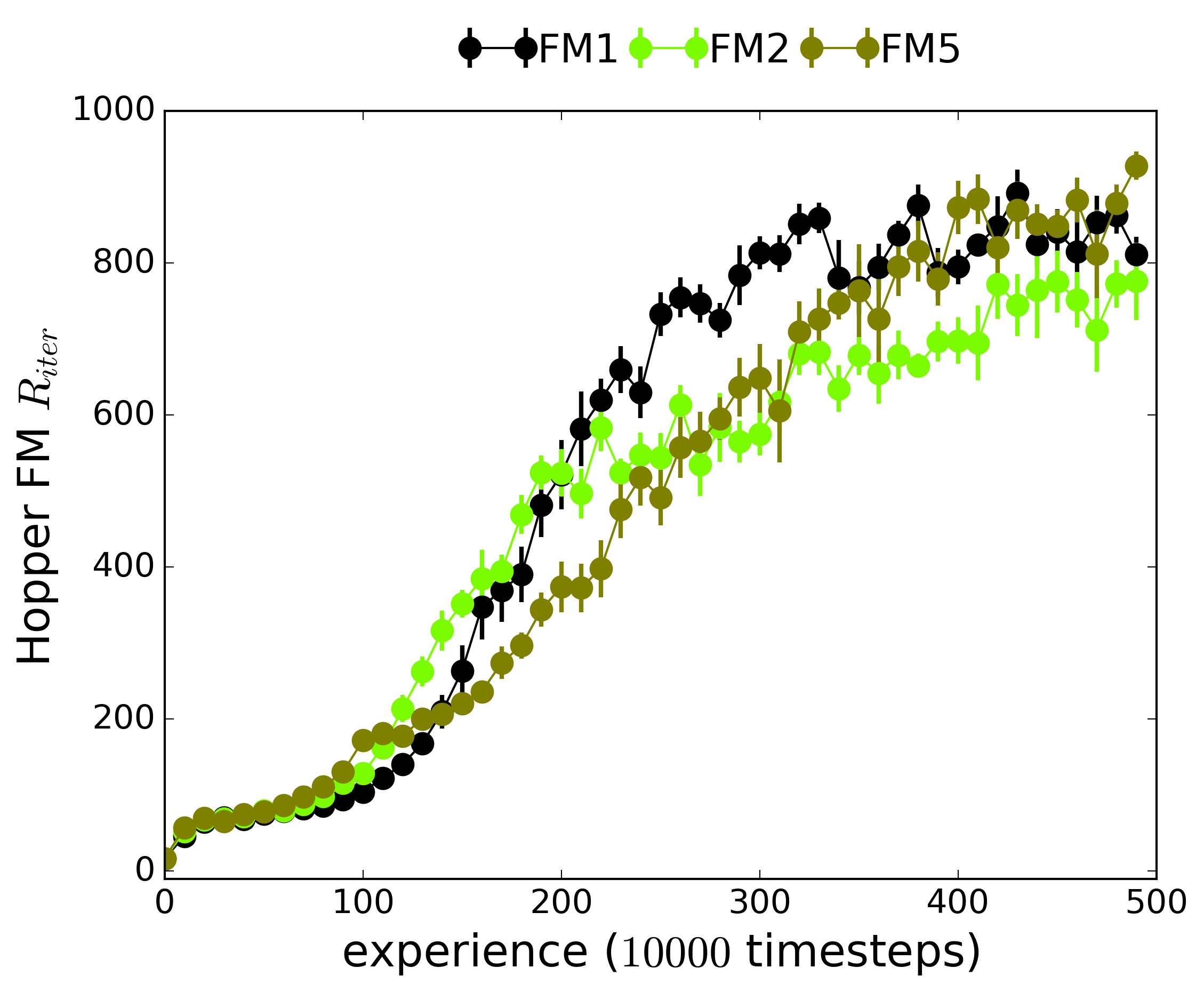}
\end{subfigure}
\begin{subfigure}[t]{0.4\textwidth}
\includegraphics[width=\textwidth]{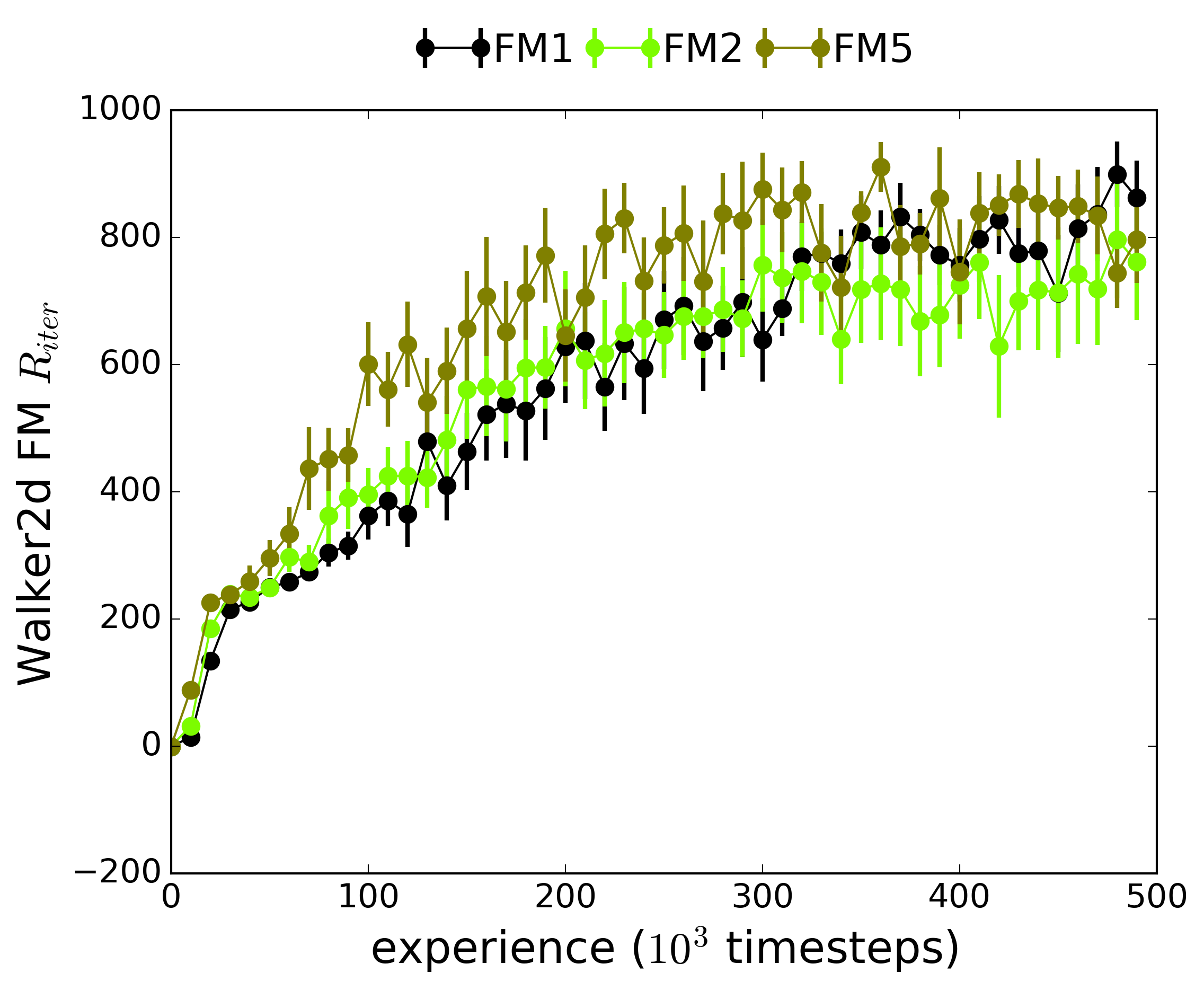}
\end{subfigure}
\begin{subfigure}[t]{0.4\textwidth}
\includegraphics[width=\textwidth]{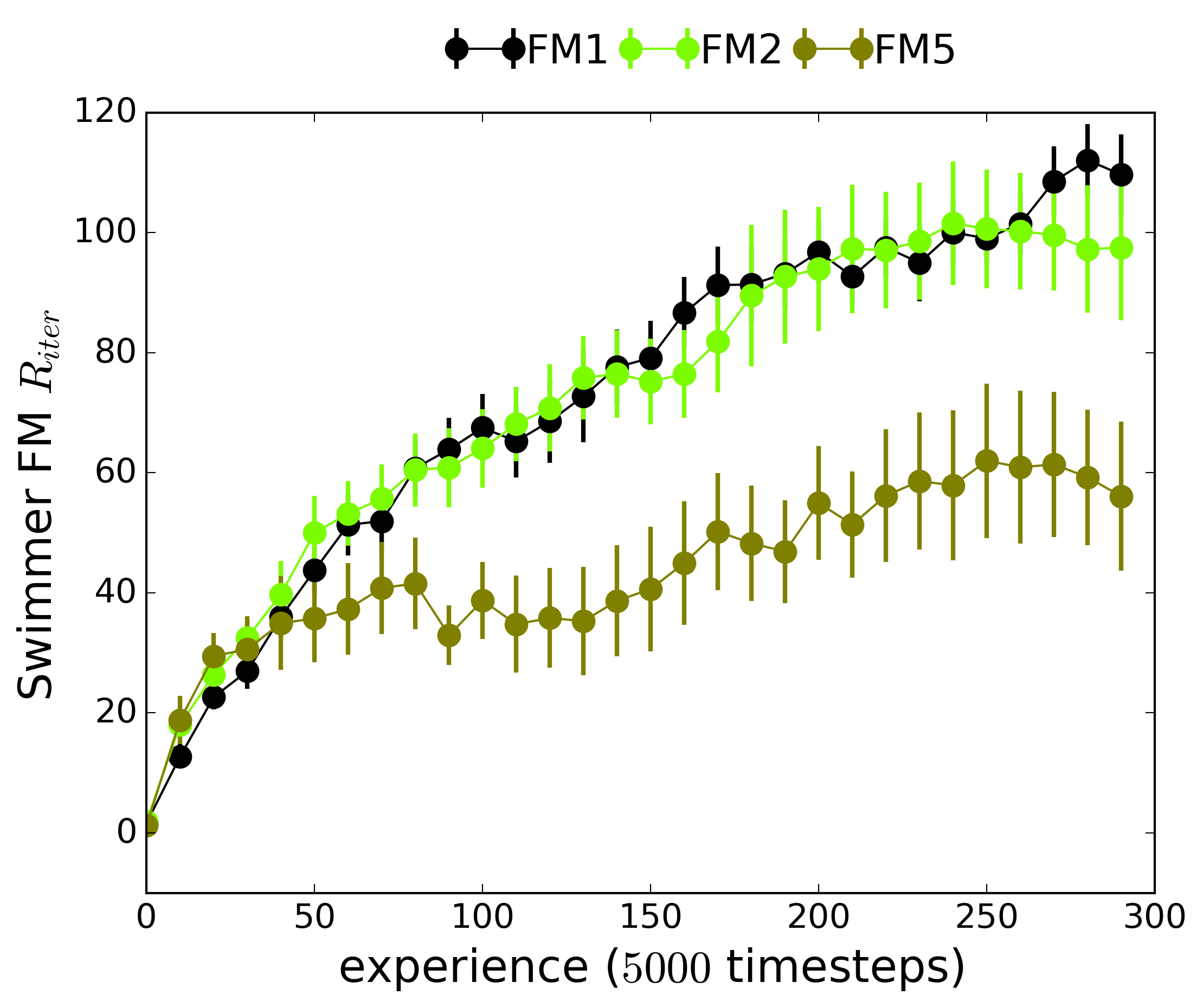}
\end{subfigure}
\caption{Empirical expected return using finite memory models of $w=1$ (black), $w=2$ (light green), $w=5$ (brown) window sizes. (top-down) Walker, Hopper, Cart-Pole, and Swimmer.}
\label{fig:FMs}
\end{figure}

\textbf{Finite Memory models:}
\label{sec:init}
Next, we present all finite memory models used as baselines in \S\ref{sec:baselines}. \figref{fig:FMs} shows finite memory models with three different window sizes $w=1,2,5$ for all environments. We report in the main comparison the best of each environment ($FM2$ for Walker, Swimmer, Cart-Pole, and $FM1$ for Hopper).

\begin{figure}[h!]
\centering
\begin{subfigure}[t]{0.4\textwidth}
\includegraphics[width=\textwidth]{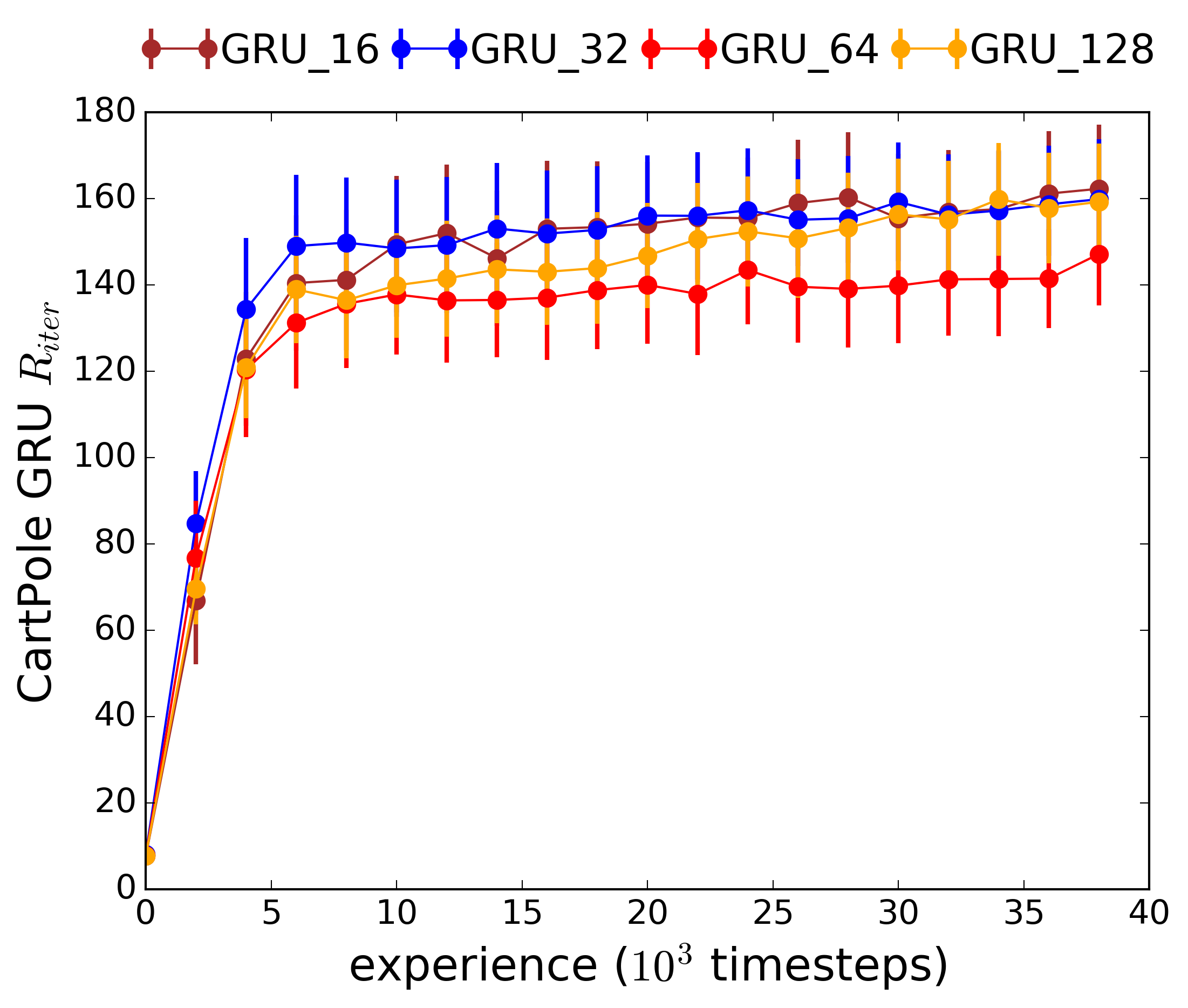}
\end{subfigure}
\begin{subfigure}[t]{0.4\textwidth}
\includegraphics[width=\textwidth]{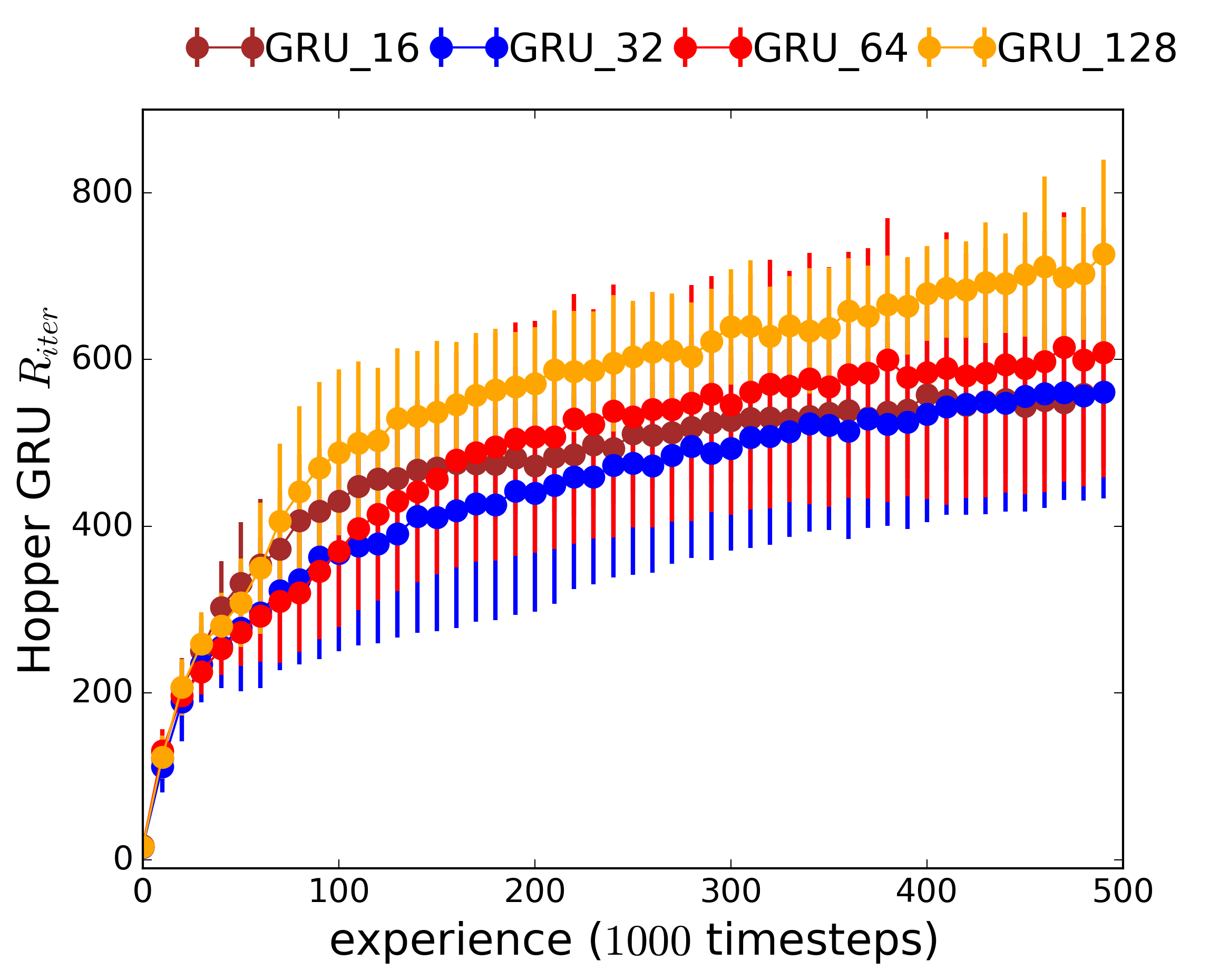}
\end{subfigure}
\begin{subfigure}[t]{0.4\textwidth}
\includegraphics[width=\textwidth]{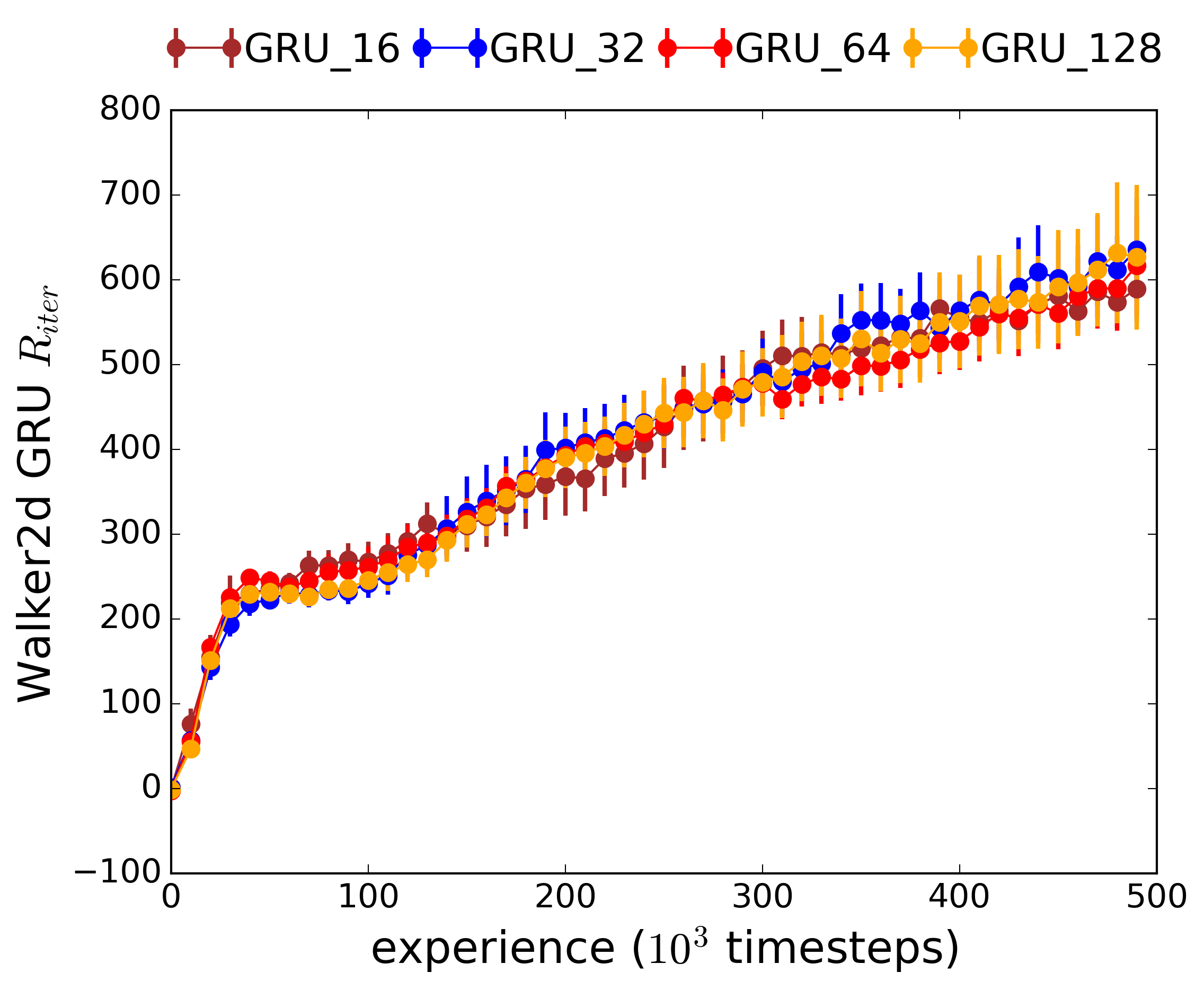}
\end{subfigure}
\begin{subfigure}[t]{0.4\textwidth}
\includegraphics[width=\textwidth]{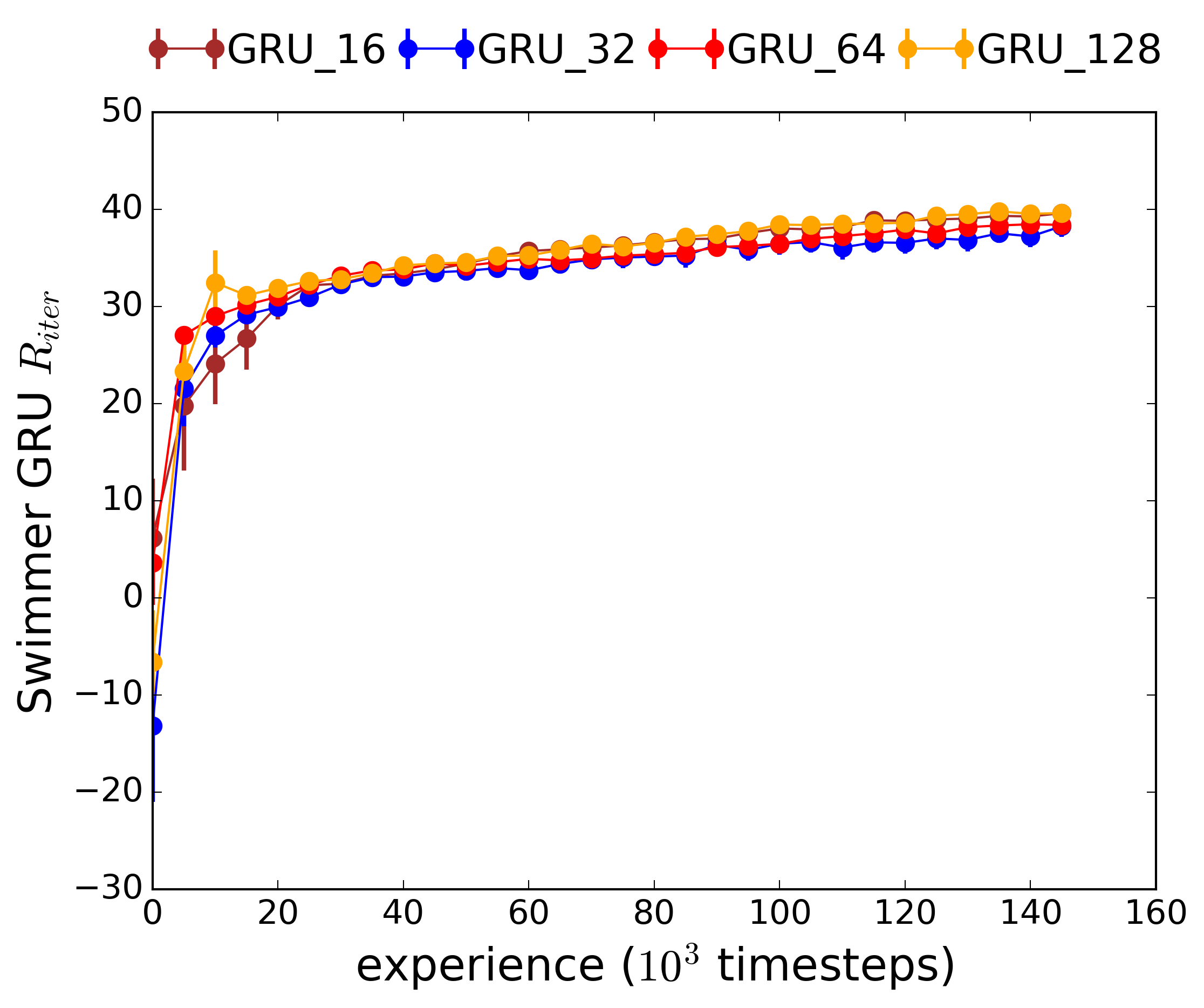}
\end{subfigure}
\caption{Empirical expected return using RNN with GRUs $d=16$ (green), $d=32$ (blue), $d=64$ (red) and $d=128$ (yellow) hidden units. (top-down) Walker, Hopper, Cart-Pole, and Swimmer.}
\label{fig:GRUs}
\end{figure}

\textbf{GRU baselines:}
\label{sec:init}
In this section we report results obtained for RNN with GRUs using the best  learning rate $\eta=0.01$.  \figref{fig:GRUs} shows the results using different number of hidden units $d=16,32,64,128$ for all the environments.

\begin{figure}[t!]
\centering
\begin{subfigure}[t]{0.43\textwidth}
\includegraphics[width=\textwidth]{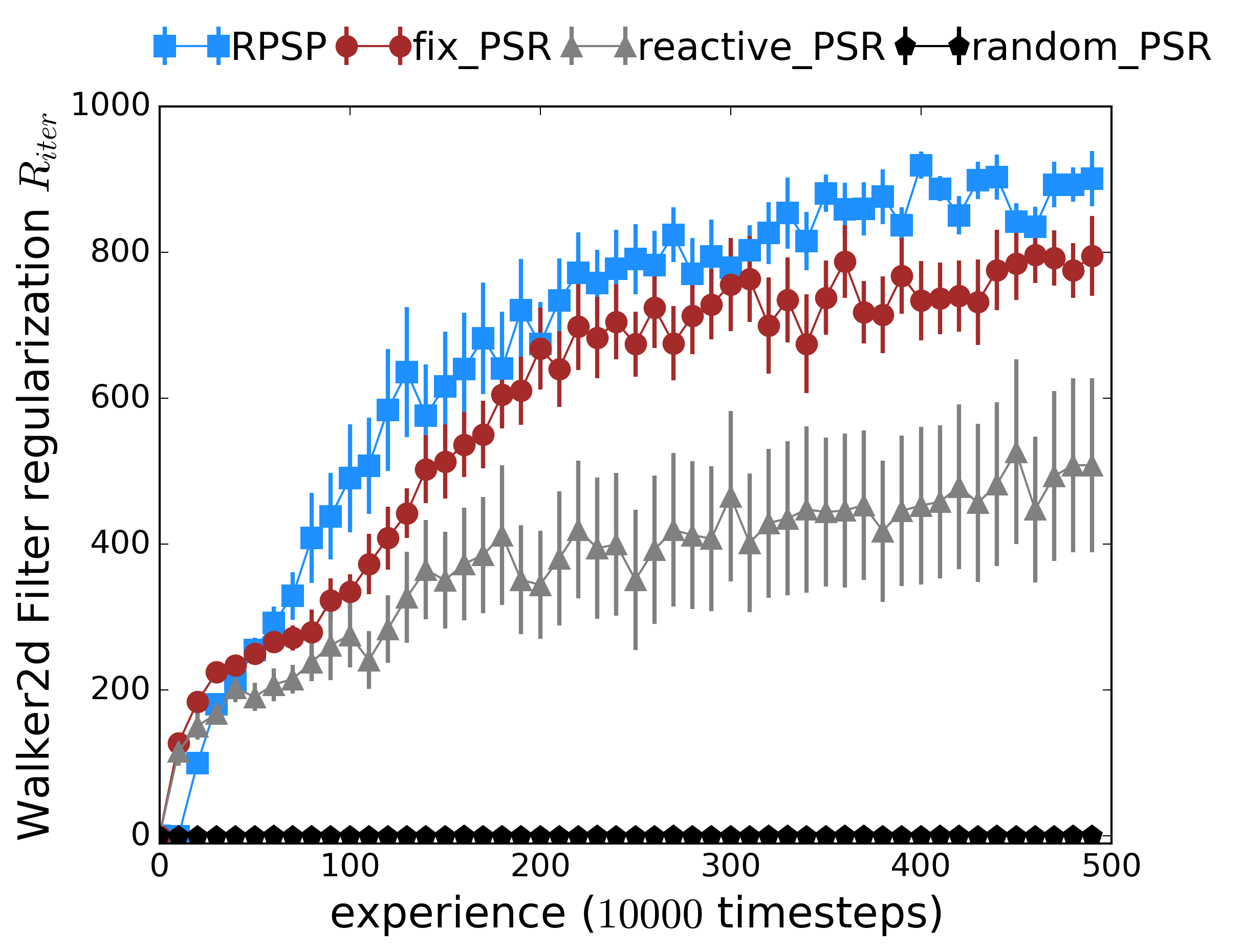}
\end{subfigure}
\begin{subfigure}[t]{0.43\textwidth}
\includegraphics[width=\textwidth]{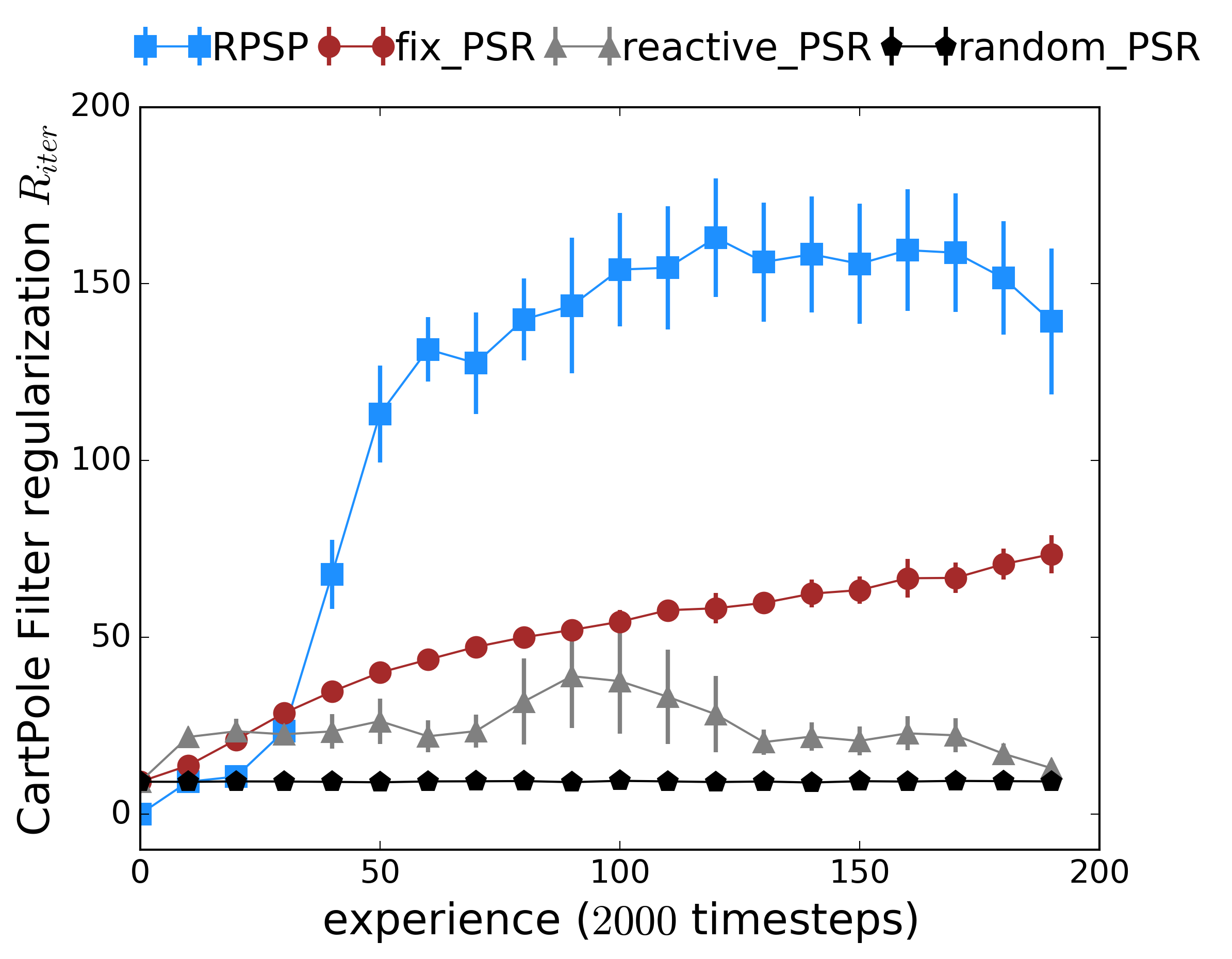}
\end{subfigure}
\begin{subfigure}[t]{0.43\textwidth}
\includegraphics[width=\textwidth]{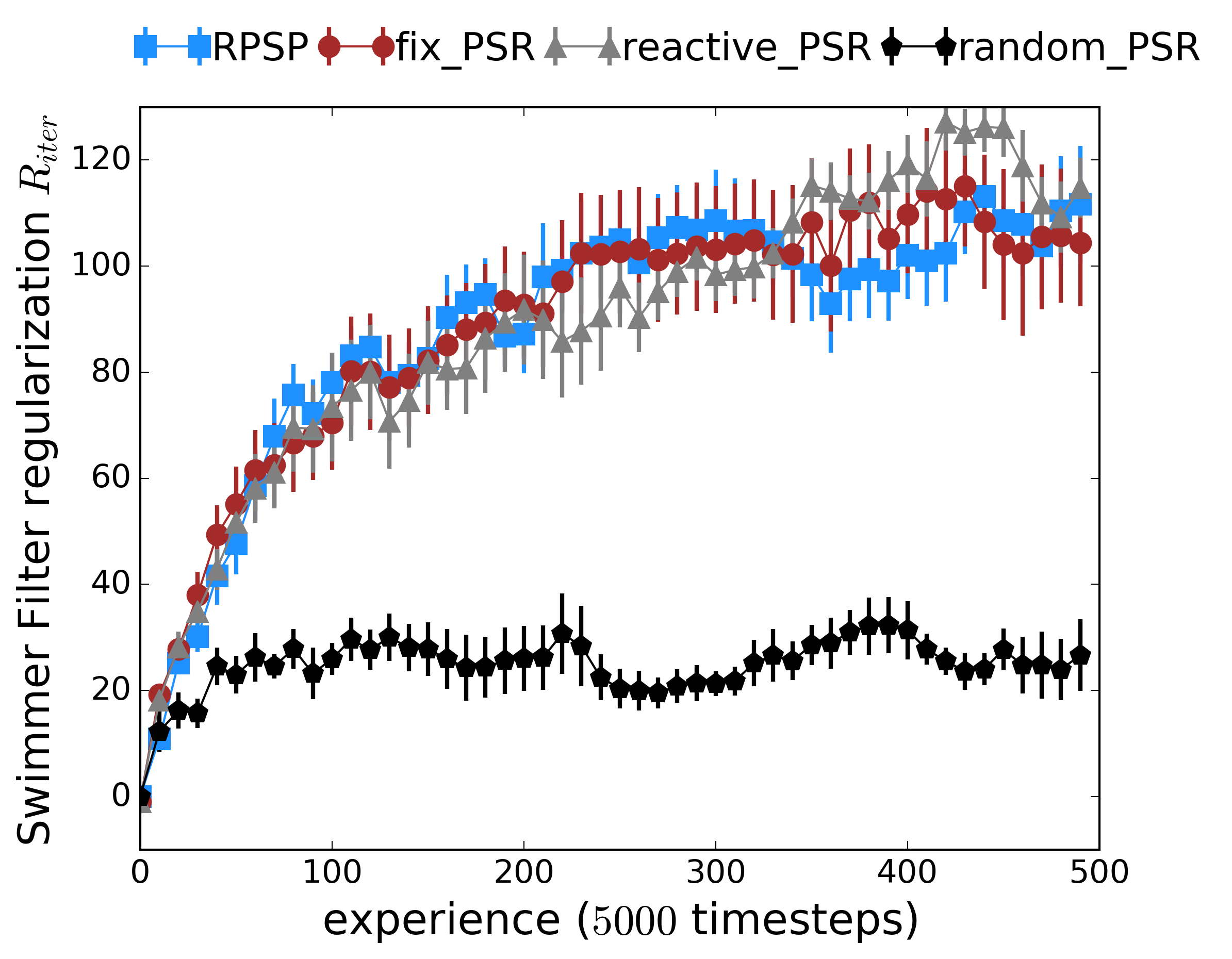}
\end{subfigure}
\caption{Predictive filter regularization effect for Walker2d, CartPole and Swimmer environments. RPSP with predictive regularization (RPSP:blue), RPSP with fixed PSR filter parameters (fix\_PSR:red), RPSP without predictive regularization loss (reactive\_PSR: grey).}
\label{fig:filterreg}
\end{figure}
\begin{figure}[t!]
\centering
\begin{subfigure}[t]{0.43\textwidth}
\includegraphics[width=\textwidth]{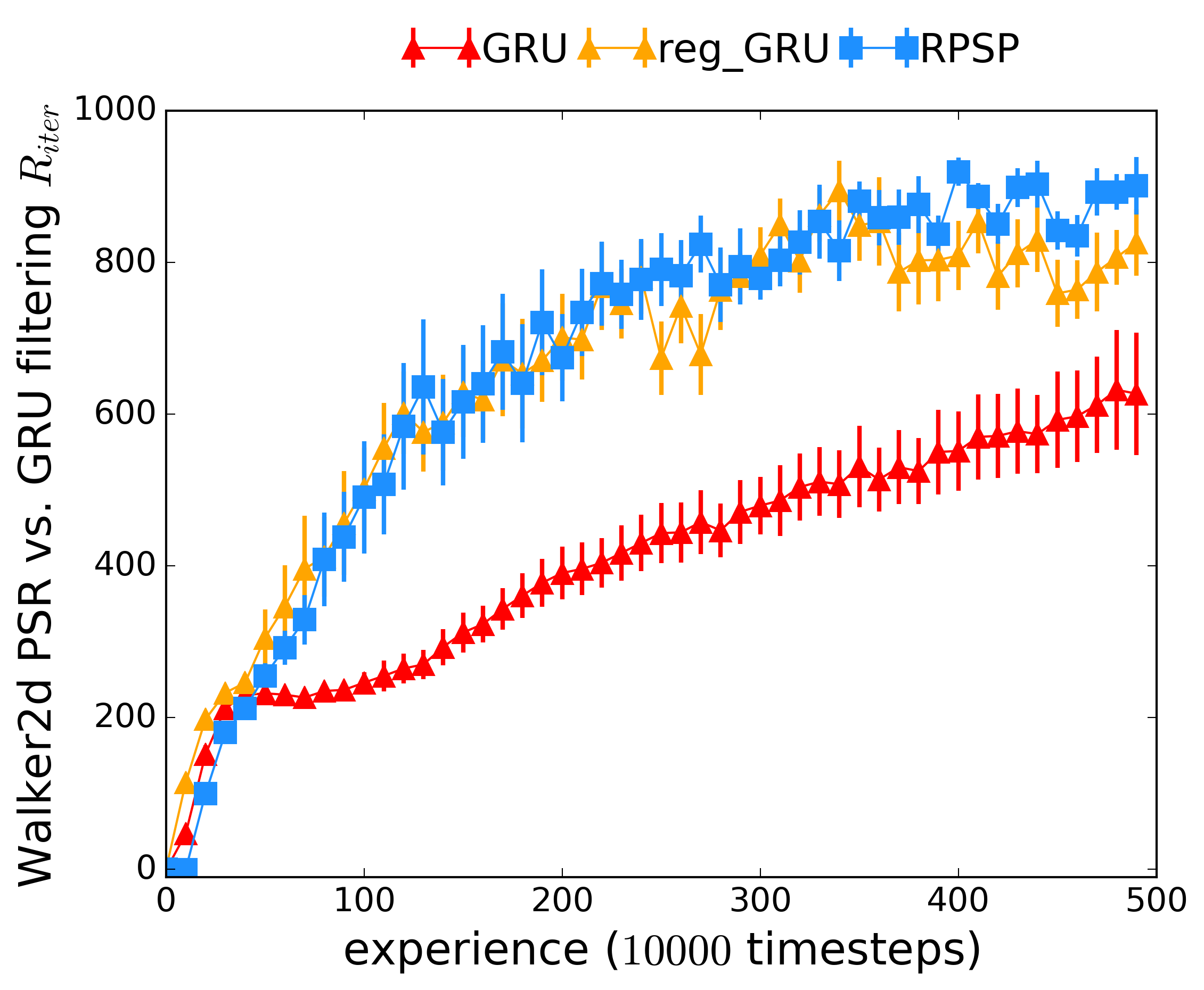}
\end{subfigure}
\begin{subfigure}[t]{0.43\textwidth}
\includegraphics[width=\textwidth]{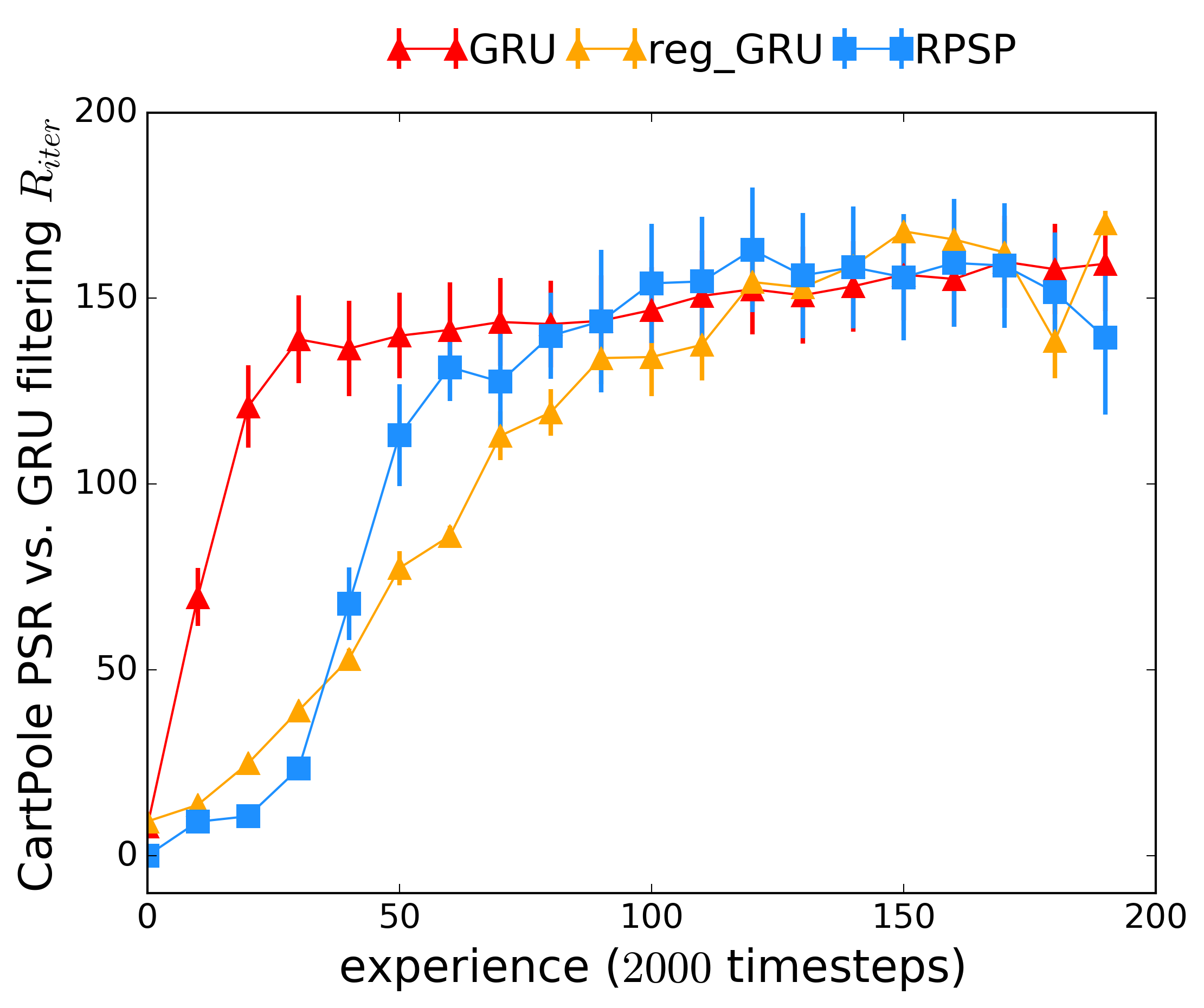}
\end{subfigure}
\begin{subfigure}[t]{0.43\textwidth}
\includegraphics[width=\textwidth]{{"figs/10of10/Swimmer_PSR_vs._GRU_filtering_rwdperiter_best10_step10"}.png}
\end{subfigure}
\caption{GRU vs. RPSP filter comparison for other Walker and CartPole environments. GRU filter without regularization loss (GRU:red), GRU filter with regularized predictive loss (reg\_GRU: yellow), RPSP (RPSP:blue)}
\label{fig:filterGRU}
\end{figure}

\clearpage

\end{document}